\documentclass[10pt,twocolumn,journal,letterpaper]{IEEEtran}
\usepackage{amsmath, amsthm, amssymb}
\usepackage{url,flushend,multirow,booktabs}
\usepackage{algorithm,algorithmic}   
\usepackage{graphicx}
\usepackage{bm}
\usepackage{cite}
\usepackage{subcaption} 
\usepackage{float}
\usepackage[dvipsnames]{xcolor}  
\usepackage{multirow}
\usepackage[colorlinks,citecolor=Green,urlcolor=blue,bookmarks=false,hypertexnames=true]{hyperref}
\usepackage{colortbl}
\definecolor{maroon}{cmyk}{0.08,0.04,0.00,0.06}  

\newcommand{\eg}{e.g.}
\newcommand{\ie}{i.e.}

\newcommand{\etal}{\textit{et al.}}

\renewcommand{\algorithmicrequire}{\textbf{Input:}}   
\renewcommand{\algorithmicensure}{\textbf{Output:}}  

\renewcommand{\b}{\color{blue}}

\DeclareFixedFont{\mf}{OT1}{ptm}{m}{n}{10pt}
\DeclareFixedFont{\mfb}{OT1}{ptm}{bx}{n}{10pt}

\begin{document}
%

\title{Channel-wise Dynamic Knowledge Distillation via Adaptive Sample Generation for Action Recognition}
%
%
%

\author{Ping~Li, Chenhao~Ping, Jie~Song, Mingli~Song  
	\thanks{P.~Li, and C.~Ping are with the School of Computer Science, Hangzhou Dianzi University, Hangzhou, China (e-mail: lpcs@hdu.edu.cn, pch@hdu.edu.cn).}
	\thanks{Jie~Song and M.~Song are with the State Key Laboratory of Blockchain and Data Security, Zhejiang University, Hangzhou, China (e-mail:sjie@zju.edu.cn, songml@zju.edu.cn).} 
}

\markboth{Draft}
{LI \MakeLowercase{\textit{et al.}}:~Channel-wise dynamic knowledge distillation via adaptive sample generation for action recognition}
%

\maketitle

\begin{abstract}
	Knowledge Distillation (KD) offers a promising yet underexplored path for compressing large action recognition models. However, existing KD methods suffer from two key limitations: 1) reliance on fixed input samples leads to suboptimal feature alignment between the frozen teacher (larger model) and the learnable student (smaller model), and 2) applying a uniform distillation strength for all channels fails to account for their varying importance in capturing distinct knowledge (\eg, motion tempo or magnitude) across training epochs. This motivates us to develop an Adaptive Sample-aware Channel-wise Dynamic (\textbf{ASCD}) KD approach, which operates in two stages. First, we use an adaptive sample generation module to create updated samples by incorporating semantics from sample gradients, which are derived by minimizing a feature loss weighted by channel centroid frequency differences at each layer. Meanwhile, crucial motion-related details are preserved by applying a Gaussian mask to frequency features. Second, we employ a channel-wise dynamic distillation module to train student on these generated samples, guided by sample gradients and feature frequencies. For efficiency, samples are updated periodically rather than per epoch. Extensive experiments on three video benchmarks (UCF101, Kinetics-400, Something-Something-v2) and two image datasets (CIFAR-100, ImageNet) demonstrate the state-of-the-art performance of our method. Code is available at \href{https://github.com/mlvccn/ASCD_KD_Action}{{\b https://github.com/mlvccn/ASCD\_KD\_Action}}.
\end{abstract}


\ifCLASSOPTIONpeerreview
\begin{center} \bfseries EDICS Category: 3-BBND \end{center}
\fi

\IEEEpeerreviewmaketitle

\section{Introduction}
\label{sec:intro}
\begin{figure}[!t]
	\centering
	\vspace{-1mm}
	\includegraphics[width=0.46\textwidth]{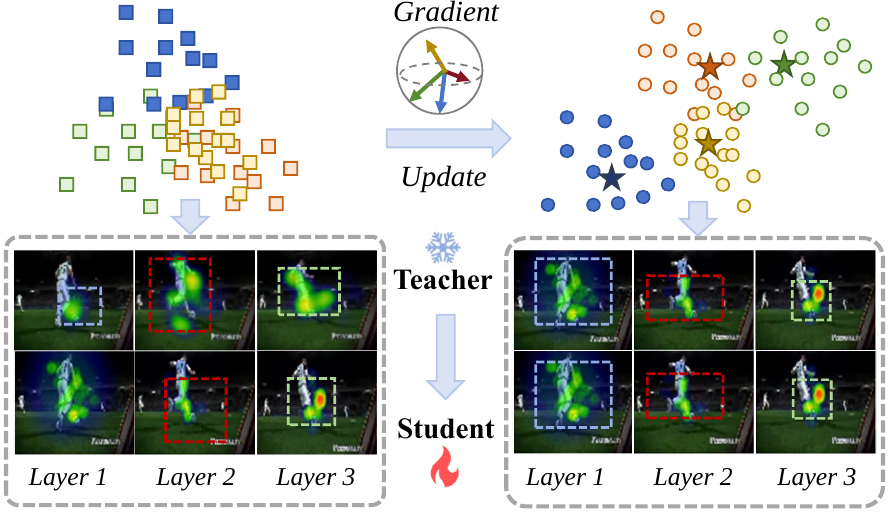}
	\caption{Motivation. Sample distribution changes while student and teacher features approach after sample update.}
	\label{fig:motivation}
	\vspace{-2mm}
\end{figure}

Action recognition has established itself as a fundamental task in computer vision, by identifying action category in video. It has wide applications spanning video surveillance, sports analytics, and autonomous driving. Early approaches leverage both RGB and optical flow features via two-stream networks \cite{simon-nips2014-twostream}, or 3D convolutions \cite{ji-tpami2012-3d} to capture spatiotemporal relations among frames. Recently, Transformer-based architectures \cite{arnab-iccv2021-vivit,yan-cvpr2022-Multiview} have achieved state-of-the-art performance by capturing long-range dependencies. However, their high computational and memory costs hinder deployment on resource-constrained devices. Model compression thus becomes essential, and Knowledge Distillation (KD) \cite{geoffrey-arxiv2015-kd, pham-wacv2024-frequency, zong-iclr2023-kd} presents a compelling technique by training a compact student network (small) to mimic a powerful teacher (large). This work focuses on addressing two fundamental challenges in KD, \ie, \textit{achieving effective feature alignment between teacher and student with dynamic inputs}, and \textit{adapting to the evolving knowledge learnt by student throughout training}. Our core motivation is illustrated in Fig.~\ref{fig:motivation}, where left(square)/right(circle) scatter plots denote that before/after sample update.

Existing methods often struggle with the inherent teacher-student capacity gap. While some simplify teacher features \cite{shu-iccv2021-channelkd, wang-nips2020-fusion} to facilitate learning, recent efforts aim to bridge this gap directly. These works include using assistant network that mixes teacher-student layers to produce consistent outputs \cite{liu-iclr2023-fkd}, transferring attention feature semantics with the generative model \cite{wang-aaai2024-gkd}, exchanging teacher-student features with cross-head prediction \cite{wang-cvpr2024-crosskd}, or embedding spatial features of heterogeneous networks to the same space \cite{guo-tpami2024-pixelkd}. However, they suffer from a limitation, \ie, relying on a static data distribution from fixed samples and a frozen teacher. This raises the \textit{first} technical problem, \ie, \textit{reliance on fixed inputs lead to suboptimal feature alignment between frozen teacher and learnable student}. 

Furthermore, the knowledge a student can absorb varies across channels and training epochs, and the dynamic aspect is largely overlooked by previous works \cite{liu-iclr2023-fkd, wang-aaai2024-gkd, wang-cvpr2024-crosskd, wang-tip2024-dkd}, which treat all layers and channels equally. Actually, different channels encode distinct semantics such as motion magnitude, action tempos, and visual cues. To this end, only a few works has explored this issue in image classification and object detection. For example, identify distillation strength of each layer using attention weights of student \cite{passban-aaai2021-alpkd} or compute channel weights by the global pooling of teacher feature map \cite{dai-aaai2025-channelkd} to obtain attention scores. However, they either neglect channel difference or use fixed channel weights, failing to adapt to varying knowledge across epochs. This raises the \textit{second} technical problem, \ie, \textit{applying a uniform distillation strength for all channels fails to account for their varying importance in capturing distinct knowledge across epochs}. 

\begin{figure*}[!t]
	\centering
	\vspace{-2mm}
	\includegraphics[width=0.85\textwidth]{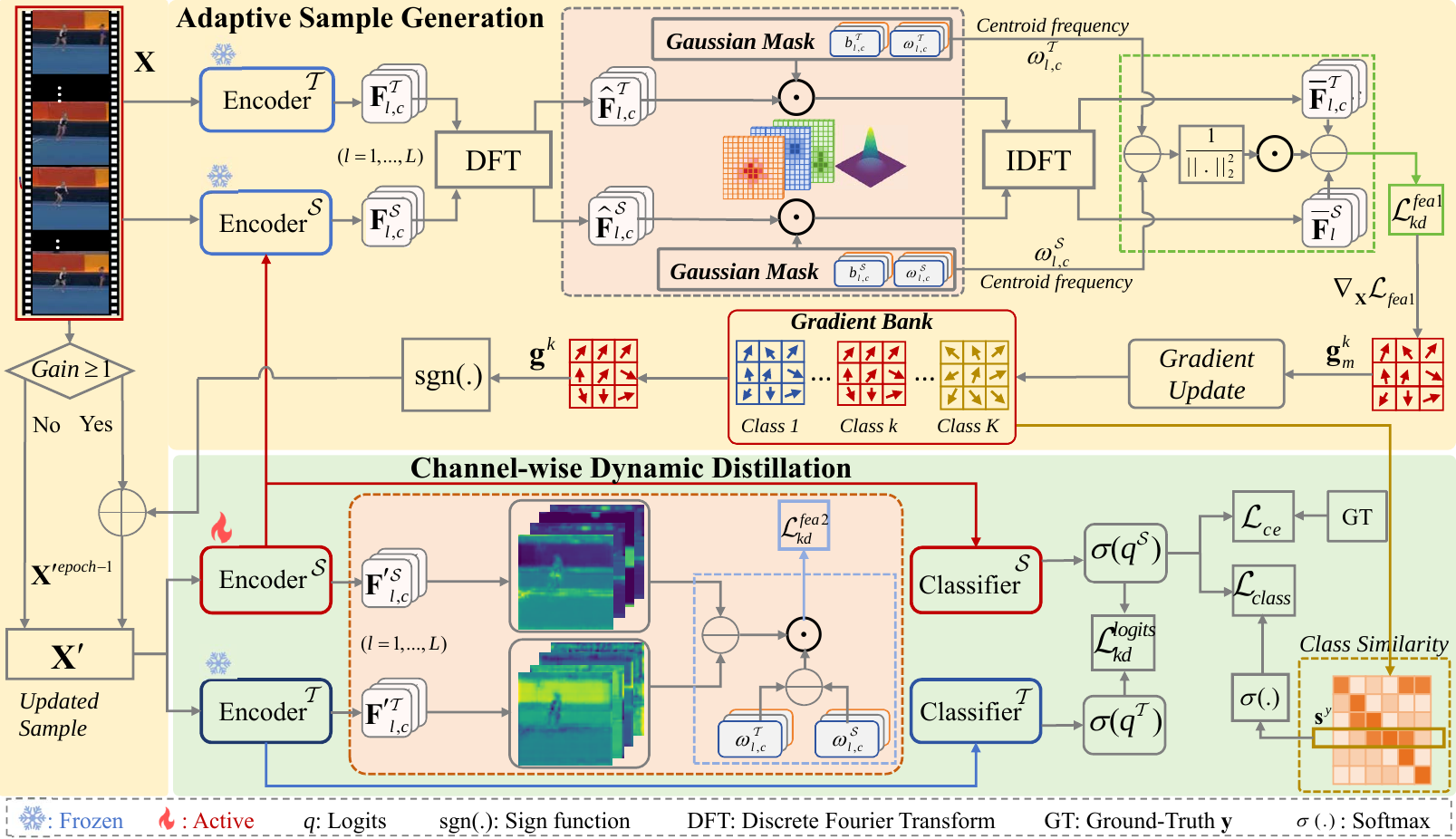} \vspace{2mm}
	\caption{Overall framework of our Adaptive Sample-aware Channel-wise Dynamic KD approach.}
	\label{fig:framework}
	\vspace{-1mm}
\end{figure*}

To address these limitations, we propose an Adaptive Sample-aware Channel-wise Dynamic (\textbf{ASCD}) KD framework for action recognition. As shown in Fig.~\ref{fig:framework}, it primarily includes an Adaptive Sample Generation (ASG) module and a Channel-wise Dynamic Distillation (CDD) module. 

First, we present an Adaptive Sample Generation (ASG) module to change the input distribution, by generating the samples that well fit feature distillation at stage one. To better capture motion dynamics, we project features from the spatial to the frequency domain, where the Gaussian mask with channel-wise learnable centroid frequency and bandwidth is used to adaptively isolate intrinsic features of specific frequency (\eg, edge) while abandoning those redundancy or noise (\eg, lighting variation). We minimize the feature loss of adaptive samples weighted by the reciprocal of centroid frequency difference between teacher and student, which results in sample gradients as guidance to update samples. This reshapes the data distribution for efficient feature alignment. Note that samples are updated only when some condition (\eg, the cumulative poly-like schedule) is satisfied to reduce costs.  

Second, we design a Channel-wise Dynamic Distillation (CDD) module to train student at stage two guided by the channel-wise centroid frequencies and sample gradients of stage one. In particular, we conduct feature alignment at each layer adopting the channel-wise dynamic weighting across epochs, which reveals the varying ability of student learning knowledge from teacher. Here, dynamic weighting stems from the fact that samples are updated regularly at stage one, which makes channel-wise frequencies and sample gradients change to narrow down the knowledge gap. Meanwhile, sample gradients are used to capture the inter-class similarity distribution, which constrains the student outputs to make them learn more discriminative features. 

Both the adaptive sample generation and the channel-wise dynamic distillation are performed in an alternative way to achieve the optimal alignment. Empirical studies on three video benchmarks, \ie, UCF101 \cite{soomro-arxiv2012-ucf101}, Kinectcs-400 \cite{kay-arXiv2017-kinetics}, and Something-Something-v2 (Sth-Sth-v2) \cite{goyal-iccv2017-sthsth}, as well as two image datasets, \ie, CIFAR-100 \cite{wah-2011-cifar} and ImageNet \cite{krizhevsky-nips2012-imagenet}, demonstrate the superiority of our approach. 

The main contributions are highlighted in the following:
\begin{itemize}   
	\item We consider efficient feature alignment in KD from the perspective of sample generation for the first time (to our best), which actively modifies the input distribution to better facilitate the transfer of knowledge.
	
	\item We employ Gaussian mask in frequency domain to adaptively preserve salient features for loss calculation, guided by centroid frequency differences, and build a class-gradient bank to provide guidance for updating samples.
	
	\item We leverage dynamic channel-wise centroid frequencies to guide feature alignment, and utilize dynamic sample gradients to model inter-class similarity for guiding output alignment during distillation.
	
\end{itemize}

\section{Related Work}
\label{sec:relate}
This section mainly discusses the most relevant works including action recognition and knowledge distillation.
\subsection{Action Recognition} 
Action recognition predicts the action label for a video. Early methods \cite{simon-nips2014-twostream} use optical flow or 3D convolution networks \cite{ji-tpami2012-3d} (\eg, C3D \cite{tran-cvpr2015-learning} and I3D \cite{carreira-cvpr2017-kinetics}) to discover motion patterns. However, optical-flow methods struggle to capture long-term dependencies, which inspires Temporal Segment Network (TSN) \cite{wang-eccv2016-action} to randomly sample frames from a group of segments, whose predictions are aggregated. To model different action tempos, SlowFast \cite{feichtenhofer-iccv2019-slowfast} appears with a slow path to capture detailed cues and a fast path to capture rapidly changing motion; Temporal Pyramid Network (TPN) \cite{yang-cvpr2020-temporal} capture various action tempos by a feature hierarchy for the backbone. Moreover, the global modeling of Transformer \cite{vaswani-nips2017-attention} has been applied to action recognition in various forms. For example, Video Transformer Network (VTN) \cite{neimark-cvpr2021-video} adds a temporal attention encoder on top of a pre-trained ViT replacing 3D convolution to handle long sequences; ViViT \cite{arnab-iccv2021-vivit} reduces video input tokens and modifies Transformer encoder components in the spatiotemporal dimension; Video Swin Transformer \cite{liu-cvpr2022-video} introduces an inductive bias of locality to reach a speed-accuracy trade-off. However, these methods rely on large-scale pre-trained image models. This encourages the emerge of pre-training video model using masked auto-encoder \cite{tong-nips2022-videomae}, which trains a robust ViT from limited data. 


There are only a few KD methods for action recognition. Early works \cite{crasto-cvpr2019-mars, stroud-wacv2020-d3d, thoker-icip2019-crosskd} focus on cross-modal distillation by transferring knowledge from optical flow model to RGB model. However, they do not cater to Transformer-based models. From the feature alignment perspective, a generative KD framework \cite{wang-aaai2024-gkd} enables the transfer of attention-based feature semantics from teacher to student between intermediate layers via conditional variation auto-encoder. From the sample importance perspective, a sample-level adaptive KD method \cite{li-mm2025-sakd} adaptively adjusts distillation ratio at sample level, while selecting the samples with both low distillation difficulty and high diversity to train model. 

\subsection{Knowledge Distillation}
KD serves as a model compression technique to train a lightweight student model by learning knowledge from a pre-trained large teacher model. It has the following two types.

\quad\textbf{Logit-based KD}. It utilizes teacher outputs (soft labels) as supervisory to guide student learning. The early work \cite{geoffrey-arxiv2015-kd} adopts softened logits to constrain student outputs, but inter-class relations of teacher logits may be unsuitable for student. This inspires Curriculum Temperature KD \cite{li-aaai2023-ctkd} to dynamically adjust the distillation temperature for reducing the semantic gap, but it suffers from unstable convergence. Hence, a logits standardization method \cite{sun-cvpr2024-lskd} sets temperature as the weighted standard deviation of logit to improve knowledge transfer efficiency. Unlike them, Yang \etal~\cite{yang-iccv2023-unifiedkd} adapt logits to student by decomposing KD loss to normalized KD loss and customized soft labels for both target class and non-target classes; Wu \etal~\cite{wu-eccv2022-tinyvit} sparsify teacher logits and store them in advance to save memory and computations, while teacher only forwards once and soft labels are reused. To address the teacher-student discrepancy in model capacity, Peng \etal~\cite{peng-icml2024-kdwav} introduce an auxiliary variable to promote the ability of student to model predictive distribution, aligned with a predefined distribution of teacher; but its strong prior assumption limits the applicability.

\textbf{Feature-based KD}. It aligns intermediate features of both teacher and student. The early work FitNet \cite{han-nips2015-learning} as pioneer prunes redundant connection in neural networks and retrains the model. Later methods measure the similarity between teacher and student features by appearance (Euclidean distance), which is insufficient and Function-Consistent Feature Distillation (FCFD) \cite{liu-iclr2023-fkd} produces similar outputs in the later layers. For object detection, CrossKD \cite{wang-cvpr2024-crosskd} delivers intermediate features of student detection head to that of teacher, and the resulting crosshead predictions are forced to mimic teacher predictions. Moreover, some work \cite{guo-tpami2024-pixelkd} adopts an input spatial representation distillation scheme to transfer spatial knowledge from large images to student input module, allowing to adjust both network architecture and image quality with pixel distillation. Unlike traditional spatial distillation, some works focus on channel-level distillation, \eg, Wang \etal~\cite{wang-nips2020-fusion} present a parameter-free multimodal fusion framework to dynamically exchange channels of cross-modality sub-networks; Shu \etal~\cite{shu-iccv2021-channelkd} normalize the activation map of each channel to obtain a soft probability map, paying more attention to salient regions; Huang \etal~\cite{huang-tpami2025-dist} consider semantic similarity between samples and each class at intra-class level. More recently, Dai \etal~\cite{dai-aaai2025-channelkd} discover that the salient channels tend to contain obvious object regions, and introduce channel-wise spatial feature distillation with fixed weights.

\section{Method}
\label{sec:method}
This section introduces the Adaptive Sample-aware Channel-wise Dynamic (\textbf{ASCD}) KD framework in Fig.~\ref{fig:framework}. 

\subsection{Problem Definition}
Knowledge distillation achieves model compression by training a lightweight student model $\mathcal{S}(\cdot)$ who learns knowledge from a large teacher model $\mathcal{T}(\cdot)$ usually pre-trained on the large-scale dataset. 

Given a set of video sequences $\{\mathcal{V}_i\}_{i=1}^N$ which consists of $N$ videos covering $K$ categories, we randomly sample $T$ frames from each video to form a set of tensors, \ie, $\mathcal{X}=\{\mathbf{X}_i\}_{i=1}^N\in \mathbb{R}^{T \times H \times W \times 3}$, \ie, RGB frames with three channels, height $H$ and width $W$. Each video has a class label $y$, which is usually stacked into an one-hot vector $\mathbf{y}\in \mathbb{R}^K$. Training videos are fed into both frozen teacher and learnable student, which employ typical backbones such as SlowFast \cite{feichtenhofer-iccv2019-slowfast} to obtain a set of $L$-layer intermediate features, \ie, $\mathcal{F}^\mathcal{T} = \{\mathbf{F}_l^\mathcal{T} | 1\le l \le L\}$ and $\mathcal{F}^\mathcal{S} = \{\mathbf{F}_l^\mathcal{S} | 1\le l \le L\}$. Due to their dimension differences, a mapping function $f(\cdot)$ (\eg, 1$\times$1 convolution or auto-encoder with temporal bilinear interpolation) is used to align features between teacher and student with the consistent dimension, \ie, $\{\mathbf{F}_l^\mathcal{T}, \mathbf{F}_l^\mathcal{S}\}\in \mathbb{R}^{T_l \times H_l \times W_l\times C_l}$. Meanwhile, it outputs the logits $\{\mathbf{q}_i^\mathcal{T}, \mathbf{q}_i^\mathcal{S}\}\in \mathbb{R}^{K}$.  

The feature-based KD loss is defined as:
\begin{equation}
	\mathcal{L}_{kd}^{fea} = \frac{1}{N} \sum_{i=1}^N \sum_{l=1}^L\| f(\mathbf{F}^T_{i,l}) - f(\mathbf{F}^S_{i,l}) \|_2^2,
\end{equation}
where $\|\cdot\|_2$ is $\ell_2$-norm. It can be easily generalized to channel-wise KD by adding the channel index $c$.

The logit-based KD loss is defined as:
\begin{equation}
	\mathcal{L}_{kd}^{logit} = \frac{1}{N} \sum_{i=1}^N \tau^2 \mathrm{KL}(\sigma(\mathbf{q}^\mathcal{T}_i / \tau), \sigma(\mathbf{q}^\mathcal{S}_i / \tau)),
	\label{eq:loss_logit}
	\vspace{-2mm}
\end{equation}
where $\mathrm{KL}(\cdot)$ is Kullback-Leibler divergence, $\tau>0$ is a temperature coefficient to smooth the probability distribution, $\sigma(\cdot)$ is the softmax function to yield probabilities.

The action recognition loss is defined as:
\begin{equation}
	\mathcal{L}_{ce} = -\frac{1}{N} \sum_{i=1}^N \mathbf{y}_i \log \sigma(\mathbf{q}^\mathcal{S}_i).
	\label{eq:loss_ce}
\end{equation}

\subsection{Overall Framework}
ASCD includes two main components, \ie, the Adaptive Sample Generation (ASG) module and the Channel-wise Dynamic Distillation (CDD) module. At stage one, both teacher and student are frozen, while updating samples guided by sample gradients governed by learnable frequency and bandwidth in the frequency domain. At stage two, updated samples are fed into frozen teacher and learnable student, while channel-wise feature alignment is dynamically guided by channel centroid frequencies of stage one. 

\subsection{Adaptive Sample Generation}
Existing KD methods adopt the fixed inputs (static samples) to align features or logits, which limits the distillation performance since the original data distribution may be not the optimal to transfer knowledge from teacher to student. Hence, we design an Adaptive Sample Generation (ASG) module to adaptively generate samples guided by gradients to make a better feature alignment between teacher and student. The technical motivation of this module comes from two aspects. On the one side, different layers or different channels of each layer may capture diverse properties of samples, \eg, low-level visual cues (such as edges) and high-level semantics (such as motion variations). Hence, we conduct the channel-wise feature alignment across layers. On the other side, video or image has inherent redundancy, and can be reduced in the frequency domain. This inspires us to project the features from the spatial domain to the frequency domain, where the important frequency features (\eg, relevant motion signals) are preserved while those irrelevant ones (\eg, camera jitter, illumination variation) are suppressed by an adaptive Gaussian mask with learnable centroid frequency. 

The videos of $\mathcal{X}=\{\mathbf{X}_i\}_{i=1}^N\in \mathbb{R}^{T \times H \times W \times 3}$ are input to both teacher ($\mathcal{T}$) and student ($\mathcal{S}$), which are typical action recognition models, such as SlowFast \cite{feichtenhofer-iccv2019-slowfast} and VideoST \cite{liu-cvpr2022-video}, to obtain $L$-layer intermediate features. Both teacher and student features are made to have the same dimension by adopting $1\times 1$ spatial convolution and temporal interpolation, \ie, $\{\mathbf{F}_l^\mathcal{T}, \mathbf{F}_l^\mathcal{S}\}\in \mathbb{R}^{T_l \times H_l \times W_l\times C_l}$, $l=1, \cdots, L$. Thus, the features of each channel are $\{\mathbf{F}_{l,c}^\mathcal{T}, \mathbf{F}_{l,c}^\mathcal{S}\}\in \mathbb{R}^{T_l \times H_l \times W_l}$.

To enhance spatiotemporal modeling, we project video features from the spatial domain to the frequency domain by 3D Discrete Fourier Transform (DFT). Their spatial frequencies are decomposed into low-frequency and high-frequency components, where the former represents the structure, smooth area (small changes), and semantic content (\eg, a wall or object body), while the latter reveals finer motion details, edges, and noise (\eg, texture, sharp contours). Meanwhile, their temporal frequencies are decomposed into low-temporal and high-temporal components, where the former indicates slow, gradual changes or static background, while the latter reveals fast motion and rapid changes. Moreover, DFT captures long-range dependencies by global operation and enables fast processing via multiplication. In particular, the frequency feature of channel $c$ at layer $l$ is calculated by (superscripts $\mathcal{T}$, $\mathcal{S}$ omitted)
\begin{equation}
	\hat{\mathbf{F}}_{l,c}(z, u, v) = \sum_{t,h,w=0}^{\scalebox{0.5}{$T_l-1,H_l-1,W_l-1$}}
	\mathbf{F}_{l,c}(t,h,w) \cdot 
	e^{-j2\pi( \frac{zt}{T_l} + \frac{u h}{H_l} + \frac{v w}{W_l} )},
	\label{eq:dft_fea}
\end{equation}
where $z=0, 1, \cdots, T_l-1$, $u=0, 1, \cdots, H_l-1$, $v=0, 1, \cdots, W_l-1$ are the temporal and spatial frequency indices, and $j$ is an imaginary unit. 

\textbf{Gaussian Mask}. To preserve semantically salient motion patterns for a video, we introduce the Gaussian mask $\mathrm{Mask}(\cdot)$ as a filter with a dynamic centroid frequency $(z_o,u_o,v_o)$ and its bandwidth $b_{l,c}$, which are learnable parameters by back-propagation, \ie,
\begin{equation}
	\mathrm{Mask}_{l,c}(z,u,v) = 
	e^{ - \frac{(z-z_o)^2+(u-u_o)^2+(v-v_o)^2}{2 b_{l,c}^2}}.
	\label{eq:mask}
\end{equation}
The Gaussian mask around the centroid $\omega_{l,c} = (z_o,u_o,v_o)$ preserves a band of related frequencies, \ie, fundamental frequency of the action and its immediate harmonics. This corresponds to the coherent motion pattern across different spatial regions in the video.

We apply Gaussian mask onto channel-wise frequency features in Eq.~(\ref{eq:dft_fea}), resulting in filter features, \ie, 
\begin{equation}
	\tilde{\mathbf{F}}_{l,c}(z,u,v) = 
	\hat{\mathbf{F}}_{l,c}(z,u,v) \odot \mathrm{Mask}_{l,c}(z,u,v),
	\label{eq:mask_fea}
\end{equation}
where ``$\odot$'' is element-wise product. Since the centroid frequency is dynamically learned, the filter adapts to different action patterns across videos within the same class. It will preserve both the high-frequency components of a rapid action (\eg, hand clap) and the lower-frequency components of a slower action (\eg, standing). This provides an adaptive temporal smoothing tuned to the primary motion. 

Then, the filter features are projected back to the spatial domain by Inverse Discrete Fourier Transform (IDFT), \ie,
\begin{equation}
	\tilde{\mathbf{F}}_{l,c}^{-1}(t,h,w) = \frac{1}{N_f} \sum_{z,u,v=0}^{\scalebox{0.5}{$T_l-1,H_l-1,W_l-1$}}
	\tilde{\mathbf{F}}_{l,c}(z,u,v) \cdot 
	e^{j2\pi( \frac{zt}{T_l} + \frac{u h}{H_l} + \frac{v w}{W_l} )},
	\label{eq:idft_fea}
\end{equation}
where $N_f=T\cdot H\cdot W$ is the number of frequency components as normalization factor. The real parts are kept to form the enhanced features $\{\bar{\mathbf{F}}_{l,c}^\mathcal{T}, \bar{\mathbf{F}}_{l,c}^\mathcal{S}\}\in \mathbb{R}^{T_l \times H_l \times W_l}$.

Now, we compute the weighted feature loss at stage one:
\begin{equation}
	\mathcal{L}_{kd}^{fea1} = \frac{1}{N} \sum_{i=1}^{N} \sum_{l=1}^{L} \sum_{c=1}^{C_l} \frac{\| \bar{\mathbf{F}}^{\mathcal{T}}_{i,l,c} - \bar{\mathbf{F}}^{\mathcal{S}}_{i,l,c} \|_2^2}{\| \omega_{l,c}^\mathcal{T} - \omega_{l,c}^\mathcal{S}\|_2^2+\epsilon}
	\label{eq:fea_loss_stage1}
\end{equation}
where $\epsilon>0$ (set to 1e-6) is a smoothing factor to avoid zero denominator. Note that layer normalization is applied to the centroid frequency $\omega_l$ at each layer. The weight is the reciprocal of centroid frequency difference as an alignment strength. Since the input samples are updated at stage one, the alignment features may not be the optimal one especially when the frequency difference gets larger. So we encourage the channel-wise feature alignment for a small frequency difference and suppress that for a large one. This is because the similar centroid frequencies indicate good teacher-student feature alignment. 

\textbf{Gradient Bank}. To compute the sample gradient, we optimize the weighted feature loss in Eq.~(\ref{eq:fea_loss_stage1}) using updated samples across channel dimension in the current epoch. The channel-wise gradients are concatenated to form the sample gradient $\mathbf{g}_m^k\in \mathbb{R}^{T\times H\times W\times C}$, where $m=1,2,\cdots, M_k$ indexes the $M_k$ samples in the class $k$. The samples in the same class share the class gradient. Then, $\mathbf{g}_m^k$ is used to update the gradient of class $k$ in a cumulative way, \ie,
\begin{equation}
	\mathbf{g}^k \leftarrow \frac{\mathbf{g}^k + \mathbf{g}_m^k}{\|\mathbf{g}^k + \mathbf{g}_m^k \|_2}\cdot \min(\|\mathbf{g}^k\|_2, \|\mathbf{g}_m^k \|_2),
	\label{eq:grad}
\end{equation} 
where $\mathbf{g}^k$ is initialized by $\mathbf{g}_1^k$. The summation of two sample gradients are normalized to identify the gradient direction, and the norm of the smaller gradient serves as the weight to control the magnitude of gradient update. The gradients of all classes are computed as in Eq.~(\ref{eq:grad}), leading to the gradient set $\mathcal{G}=\{\mathbf{g}^1, \mathbf{g}^2, \cdots, \mathbf{g}^K\}$ which is the gradient bank. This bank will be updated across epochs. 

\textbf{Sample Generation}. Since the class gradient reveals the periodical or repeated motion pattern of one action, we generate the new sample $\mathbf{X}^\prime \in \mathbb{R}^{T\times H\times W\times C}$ with the guidance of class gradients, \ie,
\begin{equation}
	\mathbf{X}_i^\prime = \mathbf{X}_i + \lambda \cdot lr \cdot \mathrm{sgn}(\mathbf{g}^{y_i}),
	\label{eq:smp_update}
\end{equation}  
where the hyper-parameter $\lambda>0$ and the learning rate $lr$ in current epoch is used to control the magnitude of the class gradients. Here $\mathrm{sgn}(\cdot)$ is a sign function to output positive or negative sign, \ie, the sample update direction. 

At early training epochs, student model has not yet converged and the generated gradients may be unstable, which requires more sample updates. At late training epochs, it approaches the optimal feature alignment between teacher and student, so samples should not be updated frequently. Therefore, to reduce cost and ensure stable training, we adopt a cumulative poly-like schedule to decide whether updating samples or not in current epoch. Mathematically, we define a gain function, \ie,
\begin{equation}
	\mathrm{Gain}(\eta, epoch) = \eta \cdot \sum_{p=1}^{N_{epoch}} \left(1 - \frac{epoch}{N_{max}}\right)^2
	\label{eq:gain}
	\vspace{-2mm}
\end{equation}
where $\eta>0$ is a hyper-parameter to govern the magnitude, $N_{epoch}$ is the number of epochs to make the gain to reach a threshold (set to 1) while these epochs form a set indexed by $p$, and $N_{max}$ is the maximum epoch. Note that $N_{epoch}$ may vary at each sample update, so does the \textit{epoch} set.

\subsection{Channel-wise Dynamic Distillation}
Once the samples adapting to distillation are ready, we train student model with frozen teacher at stage two. Generally, the distillation knowledge varies across different layers, while a group of channels in one layer may reveal various attributes of samples, such as texture, edge, and parts of moving object, which differ in centroid frequency. Meanwhile, the gradient bank at stage one stores the class gradients, which can be used to guide the distillation. Motivated by them, we develop the Channel-wise Dynamic Distillation (CDD) module, which consists of a weighted feature loss and a class-guided distribution loss. Details are below. 

Updated video samples $\{\mathbf{X}_i^\prime\}_{i=1}^N$ are fed into both teacher and student to obtain $L$-layer intermediate features. Their features are made to have the same dimension as at stage one, and the channel-wise features are  $\{\mathbf{F}_{l,c}^{\prime\mathcal{T}}, \mathbf{F}_{l,c}^{\prime\mathcal{S}}\}\in \mathbb{R}^{T_l \times H_l \times W_l}$.Note that layer normalization is applied to layer-wise features to stable the training. We align channel-wise features with the guidance of the corresponding centroid frequency difference. Since the input samples have been optimized for feature alignment between teacher and student at stage one, the high frequency difference indicates there is more knowledge to be learned by student and vice versa. Hence, the weighted feature loss is formulated as:
\begin{equation}
	\mathcal{L}_{kd}^{fea2} = \frac{1}{N} \sum_{i=1}^{N} \sum_{l=1}^{L} \sum_{c=1}^{C_l} \|  \mathbf{F}^{\prime\mathcal{T}}_{i,l,c} - \mathbf{F}^{\prime\mathcal{S}}_{i,l,c} \|_2^2 \cdot  \| \omega_{l,c}^\mathcal{T} - \omega_{l,c}^\mathcal{S} \|_2^2,
	\label{eq:fea_loss_stage2}
\end{equation}
where the frequency differences serve as the distillation strength to promote those channels with more learnable knowledge. 

To make student capture the discriminant structure, we approximate the class semantic distribution by computing the cosine similarity between two classes, \ie, 
\begin{equation}
	s^{y_i,k} = \frac{\langle \mathbf{g}^{y_i}, \mathbf{g}^k \rangle}{\|\mathbf{g}^{y_i}\|_2 \cdot \|\mathbf{g}^k \|_2},
	\label{eq:sim}
\end{equation}
where $\langle \cdot \rangle$ is an inner product, and $k$ indexes the class in $K$ classes. This leads to an inter-class similarity matrix $\mathbf{S}^{K\times K}$, consisting of $K$ vectors, \ie, $\mathbf{s}^k\in \mathbb{R}^K$. 

Usually, those samples with similar classes yield similar logits, so we adopt the Kullback-Leibler (KL) divergence to compute the distribution difference between student logits $\mathbf{q}$ and class similarities $\mathbf{s}$. Here the logits are output by a classifier with pooling and dropout before fully-connected layer. Then we minimize the class-guided KL loss as:
\begin{equation}
	\mathcal{L}_{class} =\frac{1}{N} \sum_{i=1}^N \tau^2 \mathrm{KL}(\sigma(\mathbf{s}^{y_i}/\tau), \sigma(\mathbf{q}_i^\mathcal{S}/\tau)),
	\label{eq:class_loss}
\end{equation}
where $\tau$ is a temperature factor as in Eq.~(\ref{eq:loss_logit}). This encourages the logit distribution of student approach the class similarity distribution. Hence, the student feature may become more discriminant during feature distillation by considering the semantic structure of class similarity. 

\subsection{Loss Function}
The total loss consists of the classification loss $\mathcal{L}_{ce}$ in Eq.~(\ref{eq:loss_ce}), and the distillation loss $\mathcal{L}_{kd}$ below, \ie,
\begin{equation}
	\mathcal{L}_{kd} = \alpha\mathcal{L}_{kd}^{fea2} + \beta\mathcal{L}_{kd}^{logit} +  \gamma\mathcal{L}_{class},	
\end{equation}
where $\{\alpha, \beta, \gamma\}>0$ are hyper-parameters to control the contributions of the three terms, $\mathcal{L}_{kd}^{logit}$ is defined in Eq.~(\ref{eq:loss_logit}), and $\mathcal{L}_{class}$ is in Eq.~(\ref{eq:class_loss}). Here, $\{\alpha,\gamma\}$ range from 0 to 1, and $\beta$ is set to 0.9 as in \cite{wang-tip2024-dkd}. So the total loss is $\mathcal{L}=\mathcal{L}_{ce}+\mathcal{L}_{kd}$. Note that $\mathcal{L}_{kd}^{fea1}$ in Eq.~(\ref{eq:fea_loss_stage1}) is only used for sample generation.


\begin{table*}[!t]
	\centering
	\caption{Performance comparison on UCF101 \cite{soomro-arxiv2012-ucf101}, Kinetics-400 \cite{kay-arXiv2017-kinetics}, and Sth-Sth-V2 \cite{goyal-iccv2017-sthsth}.}
	\label{tbl:res_video}
	\scalebox{0.88}{
		\setlength{\tabcolsep}{1.5mm}{
			\begin{tabular}{l l c cc cc cc cc cc cc cc cc cc cc}
				\toprule[0.75pt]
				\multirow{3}{*}{Method} & \multirow{3}{*}{Type} &
				\multicolumn{6}{c}{UCF101} &
				\multicolumn{6}{c}{Kinetics-400} &
				\multicolumn{6}{c}{Sth-Sth-V2} \\
				\cmidrule(lr){3-8} \cmidrule(lr){9-14} \cmidrule(lr){15-20}
				& &
				\multicolumn{2}{c}{SlowFast\cite{feichtenhofer-iccv2019-slowfast}} &
				\multicolumn{2}{c}{VideoST\cite{liu-cvpr2022-video}} &
				\multicolumn{2}{c}{TPN\cite{yang-cvpr2020-temporal}} &
				\multicolumn{2}{c}{SlowFast\cite{feichtenhofer-iccv2019-slowfast}} &
				\multicolumn{2}{c}{VideoST\cite{liu-cvpr2022-video}} &
				\multicolumn{2}{c}{TPN\cite{yang-cvpr2020-temporal}} &
				\multicolumn{2}{c}{SlowFast\cite{feichtenhofer-iccv2019-slowfast}} &
				\multicolumn{2}{c}{VideoST\cite{liu-cvpr2022-video}} &
				\multicolumn{2}{c}{TPN\cite{yang-cvpr2020-temporal}} \\
				\cmidrule(lr){3-4} \cmidrule(lr){5-6} \cmidrule(lr){7-8}
				\cmidrule(lr){9-10} \cmidrule(lr){11-12} \cmidrule(lr){13-14}
				\cmidrule(lr){15-16} \cmidrule(lr){17-18} \cmidrule(lr){19-20}
				& & Top1 & Top5 & Top1 & Top5 & Top1 & Top5 & 
				Top1 & Top5 & Top1 & Top5 & Top1 & Top5 &
				Top1 & Top5 & Top1 & Top5 & Top1 & Top5 \\
				\midrule[0.5pt]
				\multirow{2}{*}{ }&  Teac.    & 
				90.23 & 97.22 & 86.99 & 97.12 & 88.95 & 96.23 &
				62.77 & 84.55 & 74.07 & 91.05 & 63.23 & 84.46 &
				47.39 & 76.20 & 58.17 & 86.28 & 55.10 & 83.92 \\
				& Stud.  & 
				83.13 & 95.72 & 84.40 & 96.17 & 80.03 & 93.83 &
				51.08 & 76.05 & 70.81 & 89.61 & 52.13 & 76.82 &
				43.96 & 71.70 & 52.15 & 82.16 & 51.30 & 80.38 \\
				\midrule[0.5pt]
				
				CKD \cite{shu-iccv2021-channelkd} & ICCV'21 &
				84.28 & 95.10 & 84.72 & 95.77 & 82.30 & 93.29 &
				55.28 & 78.29 & 72.89 & 91.26 & 57.25 & 83.27 &
				45.29 & 73.93 & 54.28 & 83.56 & 52.48 & 81.89 \\
				DKD \cite{zhao-cvpr2022-dkd} & CVPR'22 &
				85.06 & 96.09 & 84.92 & 96.83 & 82.17 & 93.98 &
				56.17 & 80.20 & 73.95 & 92.10 & 57.24 & 82.43 &
				45.82 & 74.98 & 55.01 & 84.41 & 52.69 & 82.23 \\
				CTKD \cite{li-aaai2023-ctkd} & AAAI'23 &
				84.71 & 95.82 & 85.40 & 96.32 & 82.93 & 94.28 &
				55.21 & 79.95 & 72.23 & 90.72 & 56.34 & 82.01 &
				45.01 & 74.62 & 54.07 & 83.52 & 52.19 & 81.48 \\
				GKD \cite{wang-aaai2024-gkd} & AAAI'24 &
				84.36 & 95.23 & 84.99 & 96.04 & 83.24 & 93.57 &
				55.62 & 79.81 & 73.29 & 92.01 & 55.24 & 82.32 &
				44.72 & 73.92 & 53.89 & 83.62 & 51.92 & 81.41 \\
				CrossKD \cite{wang-cvpr2024-crosskd} & CVPR'24 &
				84.39 & 95.37 & 85.86 & 96.32 & 85.35 & 96.61 &
				\underline{58.07} & 79.26 & 74.71 & 91.86 & 56.73 & 82.62 &
				45.82 & 74.67 & 54.92 & 84.37 & 52.83 & 82.39 \\
				DualKD \cite{wang-tip2024-dkd} & TIP'24 &
				84.50 & 95.62 & 85.82 & 96.32 & 83.41 & 93.24 &
				56.25 & 79.85 & 74.09 & 91.29 & 55.32 & 82.92 &
				45.31 & 74.17 & 54.28 & 83.27 & 52.79 & 81.72 \\
				DCSF \cite{dai-aaai2025-channelkd} & AAAI'25 &
				\underline{85.92} & 96.53 & 85.92 & 96.63 & \underline{85.62} & \underline{96.94} &
				57.92 & \underline{80.93} & \underline{74.62} & 92.17 & 57.83 & \underline{83.29} &
				45.67 & 75.07 & 55.32 & \underline{85.17} & 52.93 & 82.79 \\
				SAKD \cite{li-mm2025-sakd} & MM'25 &
				84.87 & \underline{96.61} & \underline{86.49} & \underline{97.02} & 84.19 & 95.42 &
				55.26 & 79.85 & 73.01 & 91.00 & \underline{57.98} & 83.13 &
				\underline{46.61} & \underline{75.27} & \underline{56.29} & 84.82 & \underline{54.02} & \underline{83.29} \\
				DIST+ \cite{huang-tpami2025-dist} & TPAMI'25 &
				85.85 & 96.58 & 86.23 & 96.83 & 85.53 & 96.82 &
				57.29 & 80.84 & 74.39 & \underline{92.18} & 57.23 & 83.05 &
				46.09 & 75.14 & 55.81 & 84.31 & 53.82 & 83.08 \\
				Ours &  &
				\textbf{88.42} & \textbf{97.80} & \textbf{88.04} & \textbf{98.07} & \textbf{87.29} & \textbf{98.29} &
				\textbf{60.23} & \textbf{82.38} & \textbf{75.83} & \textbf{93.22} & \textbf{59.28} & \textbf{84.39} &
				\textbf{47.28} & \textbf{76.03} & \textbf{57.31} & \textbf{86.87} & \textbf{54.85} & \textbf{84.28} \\
				Gain &  &
				\textcolor{red}{+2.50} & \textcolor{red}{+1.19} & \textcolor{red}{+1.55} & \textcolor{red}{+1.05} & \textcolor{red}{+1.67} & \textcolor{red}{+1.35} &
				\textcolor{red}{+2.16} & \textcolor{red}{+1.45} & \textcolor{red}{+1.21} & \textcolor{red}{+1.04} & \textcolor{red}{+1.30} & \textcolor{red}{+1.10} &
				\textcolor{red}{+0.67} & \textcolor{red}{+0.76} & \textcolor{red}{+1.02} & \textcolor{red}{+1.70} & \textcolor{red}{+0.83} & \textcolor{red}{+0.99} \\
				\bottomrule[0.75pt]
			\end{tabular}
	}}
\end{table*}

\section{Experiments}
\label{sec:exp}
All experiments were conducted on a machine with four NVIDIA RTX 3090 graphics cards, and model was compiled using PyTorch 1.11.0, Python 3.8, and CUDA 11.4. {\b Please see more details and results in the Appendix.}

\subsection{Datasets and Evaluation Metrics}
We examine the performance on three video benchmarks, \ie, UCF101 \cite{soomro-arxiv2012-ucf101}, Kinetics-400 \cite{kay-arXiv2017-kinetics}, and Sth-Sth-v2 \cite{goyal-iccv2017-sthsth}, and two image ones, \ie, CIFAR-100 \cite{wah-2011-cifar} and ImageNet \cite{krizhevsky-nips2012-imagenet}. Following \cite{zong-iclr2023-kd,xie-cvpr2023-dynamickd}, we adopt the commonly used Top-1 and Top-5 accuracy as evaluation metrics.

\subsection{Experimental Settings}
For UCF101 \cite{soomro-arxiv2012-ucf101}, the model parameters for action recognition are initialized with the model pre-trained on Kinetics-400 \cite{kay-arXiv2017-kinetics}, the sampled frames in a clip are 32. For Kinetics-400 and Sth-Sth-v2, the sampled frames in a clip are 16. The Stochastic Gradient Descent optimizer is used with a momentum of 0.9. The initial learning rate for videos is 1e-2, with a weight decay factor of 1e-4 after each epoch. For Kinetics-400 and Sth-Sth-v2, following \cite{liu-cvpr2022-video}, both the tiny version (SwinT) and the small version (SwinS) of Video Swin Transformer (Video ST) \cite{liu-cvpr2022-video} are pre-trained on ImageNet, while models are randomly initialized. For CIFAR-100 and ImageNet \cite{krizhevsky-nips2012-imagenet}, following \cite{pham-wacv2024-frequency}, it has an initial learning rate of 0.1, a decay factor of 0.1 applied at epochs 150, 180, and 210, the batch size of 256; the maximum epoch is 240 and 200, respectively. For our method, the hyper-parameters of KD loss terms are: $\alpha$=100, $\beta$=0.9, $\gamma$=0.5; $\lambda$ in sample generation is 0.9 for video and 0.1 for image; $\eta$ in gain function is 0.7 for video and 0.3 for image. The number $L$ of intermediate layers is 4 (\eg, 4 stages for ResNet). For other methods, we run the publicly available codes from original papers to report results. Since codes of DIST+ \cite{huang-tpami2025-dist} and DualKD \cite{wang-tip2024-dkd} are unavailable, we implement them by ourself.

\begin{table}[!t]
	\centering
	\caption{Performance comparison (Top-1) on CIFAR-100 \cite{wah-2011-cifar}.}
	\label{tbl:cifar}
	\scalebox{0.8}{
		\setlength{\tabcolsep}{1.2mm}{
			\begin{tabular}{llccccc}
				\toprule[0.75pt]
				\multirow{2}{*}{Method} & \multirow{2}{*}{Type}  & WRN-40-2 & WRN-40-2 & ResNet56 & ResNet110 & ResNet32x4 \\
				&        & WRN-16-2 & WRN-40-1 & ResNet20 & ResNet32~~  & ResNet8x4~~  \\
				\midrule[0.5pt]
				&     Teac.      & 75.61    & 75.61    & 72.34    & 74.31     & 79.42      \\
				&       Stud.    & 73.26    & 71.98    & 69.06    & 71.14     & 72.50      \\
				\midrule[0.5pt]
				CTKD\cite{li-aaai2023-ctkd} &    CVPR'23         & 75.45   & 73.93    & 71.19  & 73.52     &74.28   \\
				
				NORM\cite{liu-iclr2023-norm}	 &  ICLR'23  & 75.65    & 74.82    & 71.35    & 73.67   & 76.49		\\
				CAT-KD\cite{guo-cvpr2023-catbkd}	 &  CVPR'23  & 75.60    & 74.82    & 71.62    & 73.62   & 76.61		\\
				CrossKD  \cite{wang-cvpr2024-crosskd}  &CVPR'24    &74.28    &73.87     &69.95    &72.17&74.30\\
				FAM-KD \cite{pham-wacv2024-frequency}&  WACV'24  &76.03    & \underline{74.88}    & 72.03    & 74.03     & 76.24      \\
				LS\cite{sun-cvpr2024-lskd}  &CVPR'24			&\underline{76.11}   &  74.37    & 71.43  & \underline{74.17}   & 76.62  \\
				DCSF\cite{dai-aaai2025-channelkd} &  AAAI'25 & 75.98   & 74.27    & \underline{72.05}  &73.72    &\underline{76.65}    \\
				SAKD\cite{li-mm2025-sakd} & MM'25 & 74.44  & 72.86  & 70.62 & 73.36 & 72.85 \\
				DIST+\cite{huang-tpami2025-dist}&TPAMI'25&76.09&74.82&72.01&73.98&76.19\\
				Ours   &           &  \bf{76.73}    &   \bf{75.64}     &  \bf{72.31}   & \bf{74.54}   &    \bf{77.95}   \\
				Gain   &           & \textcolor{red}{+0.62}    &   \textcolor{red}{+0.76}      &  \textcolor{red}{+0.26}   &  \textcolor{red}{+0.37}  &    \textcolor{red}{+1.30}   \\
				\toprule[0.75pt]
			\end{tabular}%
		}
	}
\end{table}

\begin{table}[!t]
	\centering
	\caption{Performance comparison on ImageNet \cite{krizhevsky-nips2012-imagenet}.}
	\label{tbl:imagenet}
	\scalebox{0.9}{
		\setlength{\tabcolsep}{2.5mm}{
			\begin{tabular}{ll cc cc}
				\toprule[0.75pt]
				\multirow{2}{*}{Method} & \multirow{2}{*}{Type} & 
				\multicolumn{2}{c}{Res34/Res18} & 
				\multicolumn{2}{c}{Res50/MNv2} \\
				\cmidrule(lr){3-4}	\cmidrule(lr){5-6}
				& & Top-1$\uparrow$ & Top-5$\uparrow$ & Top-1$\uparrow$ & Top-5$\uparrow$ \\
				\midrule[0.5pt]
				& Teac. & 73.31 & 91.42 & 76.16 & 92.86 \\
				& Stud. & 69.75 & 89.07 & 68.87 & 88.76 \\
				\midrule[0.5pt]
				CKD \cite{shu-iccv2021-channelkd} &ICCV'21&  71.25 & 90.38 & 71.42 & 90.28 \\
				DKD \cite{zhao-cvpr2022-dkd} & CVPR'22 & 71.70 & 90.41 & \underline{72.05} & \underline{91.05} \\
				CTKD \cite{li-aaai2023-ctkd} & AAAI'23 & 71.38 & 90.27 & 71.16 & 90.11 \\
				GKD \cite{wang-aaai2024-gkd} & AAAI'24 & 69.82 & 89.02 & 69.72 & 89.65 \\
				CrossKD \cite{wang-cvpr2024-crosskd} & CVPR'24 & 70.12 & 89.95 & 71.20 & 90.42 \\
				DualKD \cite{wang-tip2024-dkd} & TIP'24 & 68.92 & 88.75 & 69.28 & 89.10 \\
				DCSF \cite{dai-aaai2025-channelkd} & AAAI'25 & 71.68 & 90.58 &71.28 & 90.82 \\
				SAKD \cite{li-mm2025-sakd} & MM'25 & 71.07  & \underline{90.92}  & 71.27 & 90.73  \\
				DIST+ \cite{huang-tpami2025-dist}&TPAMI'25&\underline{72.07}&90.42&72.01&90.83\\
				Ours &   & \textbf{72.72} & \textbf{91.30} &\textbf{73.13} & \textbf{92.09}\\
				Gain &   & \textcolor{red}{+0.65} &\textcolor{red}{+0.38} & \textcolor{red}{+1.08} & \textcolor{red}{+1.04}\\ 
				\toprule[0.75pt]
			\end{tabular}
		}	
	}
	\vspace{-5mm}
\end{table}

\subsection{Quantitative Results}
We report the results in Table~\ref{tbl:res_video} for UCF101~\cite{soomro-arxiv2012-ucf101}, Sth-Sth-v2~\cite{goyal-iccv2017-sthsth}, and Kinetics-400~\cite{kay-arXiv2017-kinetics}, while that in Tables~\ref{tbl:cifar} and \ref{tbl:imagenet} for CIFAR-100~\cite{wah-2011-cifar} and ImageNet~\cite{krizhevsky-nips2012-imagenet}, respectively. The best record is in bold and the second best is underlined.

From Table~\ref{tbl:res_video}, our ASCD method consistently outperforms the competitive alternatives across three typical action recognition models on all video datasets. For example, ours has a Top-1 gain of 2.50\% and 2.16\% compared to the best candidate DCSF \cite{dai-aaai2025-channelkd} and CrossKD \cite{wang-cvpr2024-crosskd} using SlowFast \cite{feichtenhofer-iccv2019-slowfast} on UCF101 and Kinetics-400, respectively. We attribute this to the fact that we adaptively generate samples with the guidance of the centroid frequencies of both teacher and student, which encourages better feature alignment compared to those using fixed samples. Meanwhile, some methods like GKD \cite{wang-aaai2024-gkd} and DualKD \cite{wang-tip2024-dkd} equally treat all layers, leading to inferior performance; DIST+ \cite{huang-tpami2025-dist} consider layer-wise distillation weighted by feature Pearson similarity, but neglect channel difference; some others consider channel-wise distillation, \eg, CKD \cite{shu-iccv2021-channelkd} treat channels equally and align channel features without considering layer difference, and DCSF \cite{dai-aaai2025-channelkd} use fixed attention scores of teacher to weight channels neglecting channel dynamics. Unlike them, we employ the learnable centroid frequencies of teacher and student to dynamically guide the feature alignment with updating samples, which allows different channels to capture varying semantics related to motions. 

Similar observations are found in Tables~\ref{tbl:cifar} and \ref{tbl:imagenet} for image datasets. Compared to several SOTAs, our gains are generally satisfying in terms of relative improvements, \eg, it achieves a Top-1 gain of 1.30\% using ResNet32x4/8x4 on CIFAR-100 \cite{wah-2011-cifar} and a Top-1 gain of 1.08\% using ResNet50/MobileNetv2 on ImageNet \cite{krizhevsky-nips2012-imagenet}. This verifies the good generalization ability of our method from the video to image domain.

\begin{table*}[!t]
	\centering
	\caption{Ablations of individual components on five benchmarks.}
	\scalebox{0.88}{
		\setlength{\tabcolsep}{1.2mm}{
			\begin{tabular}{l cc cc cc cc cc cc cc cc c cc cc}
				\toprule[0.75pt]
				\multirow{3}{*}{ASG} & \multirow{3}{*}{CDD} 
				& \multicolumn{4}{c}{UCF101\cite{soomro-arxiv2012-ucf101}} 
				& \multicolumn{4}{c}{Kinetics-400 \cite{kay-arXiv2017-kinetics}} 
				& \multicolumn{4}{c}{Sth-Sth-v2 \cite{goyal-iccv2017-sthsth}} 
				& \multicolumn{4}{c}{CIFAR-100 \cite{wah-2011-cifar}} 
				& \multicolumn{4}{c}{ImageNet \cite{krizhevsky-nips2012-imagenet}} \\
				\cmidrule(lr){3-6} \cmidrule(lr){7-10} \cmidrule(lr){11-14} \cmidrule(lr){15-18} \cmidrule(lr){19-22}
				& & \multicolumn{2}{c}{SlowFast} & \multicolumn{2}{c}{VideoST} 
				& \multicolumn{2}{c}{SlowFast} & \multicolumn{2}{c}{VideoST} 
				& \multicolumn{2}{c}{SlowFast} & \multicolumn{2}{c}{VideoST} 
				& \multicolumn{2}{c}{ResNet} & \multicolumn{2}{c}{WRN} 
				& \multicolumn{2}{c}{ResNet} & \multicolumn{2}{c}{MNv2} \\
				\cmidrule(lr){3-4} \cmidrule(lr){5-6} \cmidrule(lr){7-8} \cmidrule(lr){9-10} 
				\cmidrule(lr){11-12} \cmidrule(lr){13-14} \cmidrule(lr){15-16} \cmidrule(lr){17-18} 
				\cmidrule(lr){19-20} \cmidrule(lr){21-22}
				& & Top1$\uparrow$ & Top5$\uparrow$ & Top1$\uparrow$ & Top5$\uparrow$ 
				& Top1$\uparrow$ & Top5$\uparrow$ & Top1$\uparrow$ & Top5$\uparrow$ 
				& Top1$\uparrow$ & Top5$\uparrow$ & Top1$\uparrow$ & Top5$\uparrow$ 
				& Top1$\uparrow$ & Top5$\uparrow$ & Top1$\uparrow$ & Top5$\uparrow$ 
				& Top1$\uparrow$ & Top5$\uparrow$ & Top1$\uparrow$ & Top5$\uparrow$ \\
				\midrule[0.5pt]
				& & 84.82 & 95.73 & 85.80 & 96.48 
				& 55.23 & 80.27 & 72.31 & 90.83 
				& 45.31 & 73.82 & 54.27 & 83.82 
				& 74.93 & 91.23 & 74.74 & 93.85 
				& 71.23 & 90.15 & 70.63 & 89.26 \\
				\checkmark & & \underline{86.90} & \underline{96.98} & 86.32 & 97.41 
				& 57.72 & \underline{81.62} & \underline{74.28} & 92.25 
				& 46.27 & 75.59 & 56.27 & 85.62 
				& 76.10 & 93.61 & 75.04 & 94.25 
				& 72.03 & 91.02 & 72.37 & 91.37 \\
				& \checkmark & 86.54 & 96.72 & \underline{86.79} & \underline{97.35} 
				& \underline{58.39} & 81.29 & 73.89 & \underline{92.89} 
				& \underline{46.83} & \underline{75.93} & \underline{56.93} & \underline{86.41} 
				& \underline{76.73} & \underline{93.71} & \underline{76.46} & \underline{94.68} 
				& \underline{72.38} & \underline{91.28} & \underline{72.89} & \underline{91.85} \\
				\checkmark & \checkmark & \bf88.42 & \bf97.80 & \bf88.04 & \bf98.07 
				& \bf60.23 & \bf82.38 & \bf75.83 & \bf93.22 
				& \bf47.28 & \bf76.03 & \bf57.31 & \bf86.87 
				& \bf77.95 & \bf95.02 & \bf76.73 & \bf95.03 
				& \bf72.72 & \bf91.30 & \bf73.13 & \bf92.09 \\
				\bottomrule[0.75pt]
			\end{tabular}
		}
	}
	\label{tbl:component}
\end{table*}

\subsection{Ablation Studies}
All settings keep the same as in training unless specified.

\begin{table}[!t]
	\centering
	\caption{Ablations of $\lambda$ on UCF101\cite{soomro-arxiv2012-ucf101} and Kinetics-400\cite{kay-arXiv2017-kinetics}.}
	\small
	\scalebox{0.7}{
		\setlength{\tabcolsep}{1mm}{
			\begin{tabular}{c ccccccc cccccc}
				\toprule[0.75pt]
				\multirow{3}{*}{$\lambda$} &&  \multicolumn{5}{c}{UCF101\cite{soomro-arxiv2012-ucf101}} &&\multicolumn{5}{c}{Kinetics-400\cite{kay-arXiv2017-kinetics}} &\\
				\cmidrule(lr{0.5pt}){2-7}  \cmidrule(lr{0.5pt}){8-13}
				&&\multicolumn{2}{c}{Top-1$\uparrow$} & &\multicolumn{2}{c}{Top-5$\uparrow$} &&\multicolumn{2}{c}{Top-1$\uparrow$} & &\multicolumn{2}{c}{Top-5$\uparrow$} &  \\ 
				\cmidrule(lr{0.5pt}){2-4}  \cmidrule(lr{0.5pt}){5-7} \cmidrule(lr{0.5pt}){8-10}  \cmidrule(lr{0.5pt}){11-13}
				&& SlowFast	& VideoST && SlowFast	& VideoST && SlowFast & VideoST && SlowFast & VideoST &\\ 
				\midrule[0.5pt]
				0.0 &&86.74& 86.92 & &96.83 & 97.43 & & 58.43&72.92 & &81.38 &92.94 &\\
				0.1 &&87.22 & 86.05 && 96.82 & 96.21 & &58.54 &73.32 & &81.78 &92.97 &\\
				0.3 &&87.53 & 86.37 & & 96.83 & 96.42 & &59.27 &73.98 & &82.03 &92.97 &\\
				0.5 &&87.88 & 86.72 && 97.38 & 96.97 & &59.82 &74.53 & &82.12 &93.02 &\\
				0.7 &&88.23 & 87.07 && \underline{97.53} & \underline{97.12} & &60.03 &\underline{75.62} & &\underline{82.35} &\underline{93.18} &\\
				0.9 && \bf88.42 & \bf88.04 && \bf97.80 & \bf98.07 & &\bf60.23 &\bf75.83 & & \bf82.38&\bf93.22 &\\
				1.0 && \underline{88.25} & \underline{87.89} && 97.29 & 96.88 & & \underline{60.13}&75.53 & &82.29&93.07&\\
				\toprule[0.75pt]
			\end{tabular}
		}
	}
	\label{tbl:lambda-ucf-kinetics}
\end{table}

\begin{table}[!t]
	\centering
	\caption{Ablations of $\eta $ on UCF101\cite{soomro-arxiv2012-ucf101} and Kinetics-400\cite{kay-arXiv2017-kinetics}.}
	\small
	\scalebox{0.7}{
		\setlength{\tabcolsep}{1mm}{
			\begin{tabular}{c ccccccc cccccc}
				\toprule[0.75pt]
				\multirow{3}{*}{$\eta $} && \multicolumn{5}{c}{UCF101\cite{soomro-arxiv2012-ucf101}} &&\multicolumn{5}{c}{Kinetics-400\cite{kay-arXiv2017-kinetics}} &\\
				\cmidrule(lr{0.5pt}){2-7} \cmidrule(lr{0.5pt}){8-13}
				&&\multicolumn{2}{c}{Top-1$\uparrow$} & &\multicolumn{2}{c}{Top-5$\uparrow$} &&\multicolumn{2}{c}{Top-1$\uparrow$} & &\multicolumn{2}{c}{Top-5$\uparrow$} & \\
				\cmidrule(lr{0.5pt}){2-4} \cmidrule(lr{0.5pt}){5-7} \cmidrule(lr{0.5pt}){8-10} \cmidrule(lr{0.5pt}){11-13}
				&& SlowFast & VideoST && SlowFast & VideoST && SlowFast & VideoST && SlowFast & VideoST &\\
				\midrule[0.5pt]
				0.0 && 86.54 & 86.79 & & 96.72& 97.35 & & 58.39&72.89 & &81.29 &92.89 &\\
				0.1 && 87.12 & 87.25 & & 96.98 & 97.52 & & 58.62 &73.34 & &81.43 &93.02 &\\
				0.3 && 87.85 & 87.62 & & 97.28 & 97.75 & & 59.08&73.91 & &82.04 &93.15 &\\
				0.5 && 88.27 & 87.89 & & 97.35 & \underline{97.88} & & 59.87 &74.82& &\underline{82.36} &93.18 & \\	
				0.7 && \bf88.42 & \bf{88.04} & & \bf{97.80} & \bf98.07 & & \underline{60.23}  &\bf75.83 && \bf82.38&\underline{93.22} &\\
				0.9 && 88.03 & \underline{87.93} & & \underline{97.62} & 97.92 & & \bf60.29& \underline{75.76}& &82.03 &\bf93.27 &\\
				1.0 && 87.98 & 87.87 & & 97.55 & 97.85 & & 60.02 &75.38 & & 82.51& 93.02&\\
				\toprule[0.75pt]
			\end{tabular}
		}
	}
	\label{tbl:eta-ucf-kinetics}
	\vspace{-2mm}
\end{table}

\begin{table}[!t]
	\centering 
	\caption{Ablations of $\alpha$ on UCF101\cite{soomro-arxiv2012-ucf101} and Kinetics-400\cite{kay-arXiv2017-kinetics}.}
	\small
	\scalebox{0.7}{
		\setlength{\tabcolsep}{1mm}{
			\begin{tabular}{c ccccccc cccccc}
				\toprule[0.75pt]
				\multirow{3}{*}{$\alpha$} &&  \multicolumn{5}{c}{UCF101\cite{soomro-arxiv2012-ucf101}} &&\multicolumn{5}{c}{Kinetics-400\cite{kay-arXiv2017-kinetics}} &\\
				\cmidrule(lr{0.5pt}){2-7}  \cmidrule(lr{0.5pt}){8-13}
				&&\multicolumn{2}{c}{Top-1$\uparrow$} & &\multicolumn{2}{c}{Top-5$\uparrow$} &&\multicolumn{2}{c}{Top-1$\uparrow$} & &\multicolumn{2}{c}{Top-5$\uparrow$} &  \\ 
				\cmidrule(lr{0.5pt}){2-4}  \cmidrule(lr{0.5pt}){5-7} \cmidrule(lr{0.5pt}){8-10}  \cmidrule(lr{0.5pt}){11-13}
				&& SlowFast	& VideoST && SlowFast	& VideoST && SlowFast & VideoST && SlowFast & VideoST &\\ 
				\midrule[0.5pt]
				10	&& 87.10	& 86.52&	&96.80		&  97.53	&		& 58.27 &74.31 & &81.57 &92.53 &\\
				30	&& 87.37   &86.92&	&96.82&	97.72			&	&58.34&74.53&  &81.79&92.62	& \\	
				75	&& 88.02	&\underline{87.89}&	& 97.52					&  \underline{97.98}	&	&	59.62&74.92&	&82.03&\underline{93.08}&\\
				100	&& \bf88.42   &\bf{88.04}	&&\underline{97.80}		&\bf98.07		&	&\bf60.23&\bf75.83&	&\bf82.38&\bf93.22			 		& \\	
				150	&& \underline{88.32}&87.48	&&\bf97.83&97.94 		&	& \underline{60.13}&\underline{75.42}&	&\underline{82.27}&92.88&\\
				300	&& 87.93 & 	87.28&	&97.62		&97.73  		&	&59.83&74.68&	&81.32&92.16&\\
				500	&&87.27	&86.93	&	&96.92	&97.13&&58.36&73.92&	&80.83&91.64&\\
				\toprule[0.75pt]
			\end{tabular}
		}
	}
	\label{tbl:alpha-ucf-kinetics}
	\vspace{-2mm}
\end{table}

\begin{table}[!t]
	\centering
	\caption{Ablations of $\gamma$ on UCF101\cite{soomro-arxiv2012-ucf101} and Kinetics-400\cite{kay-arXiv2017-kinetics}.}
	\small
	\scalebox{0.7}{
		\setlength{\tabcolsep}{1mm}{
			\begin{tabular}{c ccccccc cccccc}
				\toprule[0.75pt]
				\multirow{3}{*}{$\gamma$} &&  \multicolumn{5}{c}{UCF101\cite{soomro-arxiv2012-ucf101}} &&\multicolumn{5}{c}{Kinetics-400\cite{kay-arXiv2017-kinetics}} &\\
				\cmidrule(lr{0.5pt}){2-7}  \cmidrule(lr{0.5pt}){8-13}
				&&\multicolumn{2}{c}{Top-1$\uparrow$} & &\multicolumn{2}{c}{Top-5$\uparrow$} &&\multicolumn{2}{c}{Top-1$\uparrow$} & &\multicolumn{2}{c}{Top-5$\uparrow$} &  \\ 
				\cmidrule(lr{0.5pt}){2-4}  \cmidrule(lr{0.5pt}){5-7} \cmidrule(lr{0.5pt}){8-10}  \cmidrule(lr{0.5pt}){11-13}
				&& SlowFast	& VideoST && SlowFast	& VideoST && SlowFast & VideoST && SlowFast & VideoST &\\ 
				\midrule[0.5pt]
				0.0	&& 	86.94& 86.63	&	&97.03	&  97.52	&		&57.93 & 74.32& &80.92 &92.28 &\\
				0.1	&&  87.06  &87.07	&	&97.32	&97.63		&	&58.12&74.52 & &81.23 &92.32 &\\
				0.3	&& 88.23	&\underline{87.85}&	& 97.64&97.82&	&59.34 &74.78 & &81.82 &92.87 &\\
				0.5	&& \bf88.42   &\bf{88.04}	&&\underline{97.80}	&\bf98.07		&	&\bf60.23  &\bf75.83 && \bf82.38&\bf93.22 &\\
				0.7	&& \underline{88.33}&	87.78	&	&\bf97.92&\underline{97.98} &	&\underline{59.92}& \underline{75.63}& &\underline{82.25} & \underline{93.01}&\\
				0.9	&&87.82  &	87.52	&& 96.83&97.63	 		&	&59.42&75.34 & &81.93 &92.76 &\\
				1.0	&&87.69	&87.26&		&96.27	&97.42&&59.03 &75.21 & & 81.75&92.52 &\\
				\toprule[0.75pt]
			\end{tabular}
		}
	}
	\label{tbl:gamma-ucf-kinetics}
	\vspace{-3mm}
\end{table}

\textbf{Individual Components}. We examine the ASG and CDD modules in Table~\ref{tbl:component}. Here, the baseline (row~1) only uses $\mathcal{L}_{ce}$ and $\mathcal{L}_{kd}^{logit}$. When adding ASG by considering $\mathcal{L}_{kd}^{fea1}$ and sample updating, the performance is boosted by Top-1 gain of 2.08\% using SlowFast on UCF101. When adding CDD by considering $\mathcal{L}_{kd}^{fea2}$ and $\mathcal{L}_{class}$, the performance improves by a Top-1 gain of 2.66\% using VideoST on Sth-Sth-v2. The performance achieves the best (last row) when using the two modules, which consolidates their advantages.

\textbf{$\lambda$ and $\eta$ in ASG}. We examine the affects of hyper-parameters $\lambda$ and $\eta$ in a range of 0 to 1 with seven grids, showing the results in Tables~\ref{tbl:lambda-ucf-kinetics} and \ref{tbl:eta-ucf-kinetics}, respectively. From these records, the performance of our method achieves the best when the magnitude factor $\lambda$ and the gain factor of sample update is set to 0.9 and 0.7, respectively. This indicates that larger values bring about more performance improvements than smaller ones. 

\textbf{$\alpha$ and $\gamma$ in CDD}. We examine the influences of regularization parameters $\alpha$ for feature loss and $\lambda$ for class-guided KL loss in Tables~\ref{tbl:alpha-ucf-kinetics} and \ref{tbl:gamma-ucf-kinetics}, respectively. As seen from the table, when $\alpha$ takes a large value (say 100) and $\lambda$ takes 0.5, our method has the most promising performance. This suggests the feature loss at stage dominates the feature alignment guided by the centroid frequency from stage one. Meanwhile, the inter-class similarity distribution guides the student logit learning.

\textbf{Gain Schedule in ASG}. We examine three different gain schedules in the ASG module, including Linear, Cosine, and Poly, whose results are shown in Table~\ref{tbl:gain-ucf-kinetics}. From the table, the Poly schedule exhibits the most satisfying performance among the three. Meanwhile, the Linear schedule is always the second-best candidate very close to the best results and sometimes slightly outperforming Poly on a single metric (\eg, SlowFast Top-1 on Kinetics).

\begin{table}[!t]
	\centering
	\caption{Ablations of gain schedule for sample update on UCF101 \cite{soomro-arxiv2012-ucf101} and Kinetics-400 \cite{kay-arXiv2017-kinetics}.}
	\small
	\scalebox{0.8}{
		\setlength{\tabcolsep}{0.5mm}{
			\begin{tabular}{l ccccccc cccccc}
				\toprule[0.75pt]
				\multirow{3}{*}{Method} &&  \multicolumn{5}{c}{UCF101\cite{soomro-arxiv2012-ucf101}} &&\multicolumn{5}{c}{Kinetics-400\cite{kay-arXiv2017-kinetics}} &\\
				\cmidrule(lr{0.5pt}){2-7}  \cmidrule(lr{0.5pt}){8-13}
				&&\multicolumn{2}{c}{Top-1$\uparrow$} & &\multicolumn{2}{c}{Top-5$\uparrow$} &&\multicolumn{2}{c}{Top-1$\uparrow$} & &\multicolumn{2}{c}{Top-5$\uparrow$} &  \\ 
				\cmidrule(lr{0.5pt}){2-4}  \cmidrule(lr{0.5pt}){5-7} \cmidrule(lr{0.5pt}){8-10}  \cmidrule(lr{0.5pt}){11-13}
				&& SlowFast	& VideoST && SlowFast	& VideoST && SlowFast	& VideoST && SlowFast	& VideoST &\\ 
				\midrule[0.5pt]
				Linear	&&\underline{88.02}   &	\underline{87.69}&	&\underline{97.37}	&	\underline{97.93}	&	&\underline{59.92}	&\underline{75.52} &  &	\underline{82.04}&	\underline{92.88}		& \\	
				Cosine	&& 87.73  &	86.93&	&96.86	&	97.82	&	&59.76	&75.38&  &81.94	&92.85			& \\	
				Poly	&& \bf88.42   &\bf{88.04}	&&\bf{97.80}		&\bf98.07		&	&\bf60.23  &\bf75.83 && \bf82.38&\bf93.22 &\\
				\toprule[0.75pt]
			\end{tabular}
		}
	}
	\label{tbl:gain-ucf-kinetics}
\end{table}

\textbf{Inter-class Similarity Calculation}. We examine four forms of calculating the inter-class similarity in terms of class gradient. The results are reported in Table~\ref{tbl:clasim-ucf-kinetics}, which shows our approach demonstrates strong robustness to the choice of similarity metric, as performance variations remain within a narrow margin. Meanwhile, it suggests that cosine similarity offers the most stable and generalizable formulation for inter-class similarity modeling. Kernel-based approaches (RBF and Gaussian) remain competitive, particularly in Top-5 accuracy, while Mahalanobis distance does not exhibit clear advantages, likely due to unstable covariance estimation in high-dimensional embedding spaces.

\begin{table}[!t]
	\centering
	\caption{Ablations of inter-class similarity calculation on UCF101 \cite{soomro-arxiv2012-ucf101} and Kinetics-400 \cite{kay-arXiv2017-kinetics}.}
	\small
	\scalebox{0.7}{
		\setlength{\tabcolsep}{0.63mm}{
			\begin{tabular}{l ccccccc cccccc}
				\toprule[0.75pt]
				\multirow{3}{*}{Similarity} &&  \multicolumn{5}{c}{UCF101\cite{soomro-arxiv2012-ucf101}} &&\multicolumn{5}{c}{Kinetics-400\cite{kay-arXiv2017-kinetics}} &\\
				\cmidrule(lr{0.5pt}){2-7}  \cmidrule(lr{0.5pt}){8-13}
				&&\multicolumn{2}{c}{Top-1$\uparrow$} & &\multicolumn{2}{c}{Top-5$\uparrow$} &&\multicolumn{2}{c}{Top-1$\uparrow$} & &\multicolumn{2}{c}{Top-5$\uparrow$} &  \\ 
				\cmidrule(lr{0.5pt}){2-4}  \cmidrule(lr{0.5pt}){5-7} \cmidrule(lr{0.5pt}){8-10}  \cmidrule(lr{0.5pt}){11-13}
				&& SlowFast	& VideoST && SlowFast	& VideoST && SlowFast	& VideoST && SlowFast	& VideoST &\\ 
				\midrule[0.5pt]
				Mahalanobis 	&&88.24 	& \bf88.12 &	&97.40	&	\underline{97.92}	&	&59.27	& 75.52&  &\underline{82.13}	&93.02	& \\	
				RBF Kernel  	&&\underline{88.37}   &87.92	&	&\bf98.05	&97.82		&	&59.32	&\underline{75.72} &  &	82.09&\underline{93.17}	& \\	
				Gaussian kernel	&& 88.01  &	87.38&&97.29		&97.38		&	&\underline{60.01}	&75.69 &  &82.01	&92.89	& \\	
				Cosine && \bf88.42   &\underline{88.04}	&&\underline{97.80}		&\bf98.07		&	&\bf60.23  &\bf75.83 && \bf82.38&\bf93.22 &\\
				\toprule[0.75pt]
			\end{tabular}
		}
	}
	\label{tbl:clasim-ucf-kinetics}
\end{table}

\begin{table}[!t]
	\centering
	\caption{Ablations of feature alignment strategy at stage one on UCF101 \cite{soomro-arxiv2012-ucf101} and Kinetics-400 \cite{kay-arXiv2017-kinetics}.}
	\label{tbl:feaalign-stage1-ucf-kinetics}
	\scalebox{0.8}{
		\setlength{\tabcolsep}{0.5mm}{
			\begin{tabular}{l ccccccc ccccc}
				\toprule[0.75pt]
				\multirow{3}{*}{Strategy} &&  \multicolumn{5}{c}{UCF101\cite{soomro-arxiv2012-ucf101}} &&\multicolumn{5}{c}{Kinetics-400\cite{kay-arXiv2017-kinetics}} \\
				\cmidrule(lr{0.5pt}){2-7}  \cmidrule(lr{0.5pt}){8-13}
				&&\multicolumn{2}{c}{Top-1$\uparrow$} & &\multicolumn{2}{c}{Top-5$\uparrow$} &&\multicolumn{2}{c}{Top-1$\uparrow$} & &\multicolumn{2}{c}{Top-5$\uparrow$}   \\ 
				\cmidrule(lr{0.5pt}){2-4}  \cmidrule(lr{0.5pt}){5-7} \cmidrule(lr{0.5pt}){8-10}  \cmidrule(lr{0.5pt}){11-13}
				&& SlowFast	& Video ST && SlowFast	& Video ST &&  SlowFast	& Video ST && SlowFast	& Video ST \\ 
				\midrule[0.5pt]
				
				1$\times$1 Conv 	&&85.23  &86.27&&95.89	&96.63			&&55.64	&72.81& &81.23	&91.28 \\	
				\midrule[0.5pt]
				
				Attention	&&85.38  &86.99&&96.10	&96.78	&&56.02	&73.19&  &81.89	&91.83	 \\	
				&& \textcolor{red}{+0.15} & \textcolor{red}{+0.72} && \textcolor{red}{+0.21} & \textcolor{red}{+0.15} && \textcolor{red}{+0.38} & \textcolor{red}{+0.38} && \textcolor{red}{+0.66} & \textcolor{red}{+0.55} \\
				
				Frequency 	&&87.29   &87.93&&97.31		&97.23	&	&58.32	&74.38 &  &\underline{81.93}	&91.73	 \\
				&& \textcolor{red}{+2.06} & \textcolor{red}{+1.66} && \textcolor{red}{+1.42} & \textcolor{red}{+0.60} && \textcolor{red}{+2.68} & \textcolor{red}{+1.57} && \textcolor{red}{+0.70} & \textcolor{red}{+0.45} \\
				
				Numerator&& \underline{87.38} &\bf{88.12}&&\underline{97.52}		&\underline{97.82}	&	&\underline{59.38}&\underline{75.02} &  &81.76	&\underline{92.38}	 \\
				
				&&\textcolor{red}{+2.15} & \textcolor{red}{+1.85} && \textcolor{red}{+1.63} & \textcolor{red}{+1.19} && \textcolor{red}{+3.74} & \textcolor{red}{+2.21} && \textcolor{red}{+0.53} & \textcolor{red}{+1.10} \\ 
				
				\midrule[0.5pt]
				Ours&& \bf88.42   &\underline{88.04}	&&\bf{97.80}		&\bf98.07		&	&\bf60.23  &\bf75.83 && \bf82.38&\bf93.22 \\
				&& \textcolor{red}{+3.19} & \textcolor{red}{+1.77} && \textcolor{red}{+1.91} & \textcolor{red}{+1.44} && \textcolor{red}{+4.59} & \textcolor{red}{+3.02} && \textcolor{red}{+1.15} & \textcolor{red}{+1.94} \\
				
				\toprule[0.75pt]
			\end{tabular}
		}
	}
\end{table}

\textbf{Feature Alignment Strategy at Stage One}. We adopt the weighted feature loss $\mathcal{L}_{kd}^{fea1}$ at stage one, and examine other strategies in Table~\ref{tbl:feaalign-stage1-ucf-kinetics}. Row~1 indicates the 1$\times$1 convolution without learnable Gaussian mask, on the basis of which Row~2 adds two attention layers to enhance feature alignment between teacher and student. Row~3 aligns features in frequency domain, Row~4 regards the centroid frequency difference as the weight. Compared to the baseline (Row~1), there is a large improvement of 3-4\% by our alignment form that adopts the reciprocal of the centroid frequency difference as the weight of the feature MSE loss. Frequency domain alignment (Row 3) shows substantial gains over spatial methods (Row 1\&2) while numerator weighting (Row 4) further improves upon frequency alignment. Ours adopts the reciprocal weighting which considers the fact that centroid frequencies are unstable at early training epochs which discourages the feature alignment when the frequency difference is large.  

\begin{table}[!t]
	\centering
	\caption{Ablations of heterogeneous architectures for teacher and student on UCF101 \cite{soomro-arxiv2012-ucf101}.}
	\label{tbl:cross-modal-ucf}
	\scalebox{0.9}{
		\setlength{\tabcolsep}{2.2mm}{
			\begin{tabular}{l l ccc cc}
				\toprule[0.75pt]
				\multirow{2}{*}{Method} & \multirow{2}{*}{Venue} &&
				\multicolumn{2}{c}{TPN $\rightarrow$ SlowFast} & 
				\multicolumn{2}{c}{VideoST$\rightarrow$SlowFast} \\
				\cmidrule(lr){4-5} \cmidrule(lr){6-7}
				& & & Top-1$\uparrow$ & Top-5$\uparrow$ & Top-1$\uparrow$ & Top-5$\uparrow$ \\
				\midrule[0.5pt]
				Teac. & - & & 88.95 & 96.23 & 86.99 & 97.12 \\
				\midrule[0.5pt]			
				DKD \cite{zhao-cvpr2022-dkd}& CVPR'22 & & 85.36 & 96.13 & 85.02 &97.01  \\
				CTKD\cite{li-aaai2023-ctkd} & AAAI'23 & & 84.92 & 95.89 & 84.52 & 96.33 \\
				GKD\cite{wang-aaai2024-gkd} & AAAI'24 & & 85.53 & 96.14 & 85.58 & 96.59 \\
				CrossKD \cite{wang-cvpr2024-crosskd} & CVPR'24 & & 85.92 & 96.03 & 85.92 & 96.77 \\
				DualKD \cite{wang-tip2024-dkd}& TIP'24 & & 86.22 & 96.23 & 86.08 & 96.92 \\
				DCSF	\cite{dai-aaai2025-channelkd} &AAAI'25 &&86.34& 96.53  &86.22 &96.45\\
				SAKD\cite{li-mm2025-sakd} & MM'25 && \underline{86.46} & \underline{96.71}  & \underline{86.79} & \underline{96.69}  \\
				DIST+\cite{huang-tpami2025-dist}&TPAMI'25&&86.02&96.35&85.28&96.13\\
				\midrule[0.5pt]
				Ours& &&\bf{87.45} & \bf{97.23}  &\bf{87.12 } &\bf{97.23}\\
				\toprule[0.75pt]
			\end{tabular}
		}
	} 
\end{table}

\textbf{Teacher-Student Heterogeneity}. We examine the performance when teacher and student adopt different architectures, showing the results in Table~\ref{tbl:cross-modal-ucf}. From the table, our method consistently outperforms all existing KD approaches across heterogeneous teacher-student pairs on UCF101 by achieving the best Top-1 and Top-5 accuracy. This indicates our method successfully bridges architectural gaps between different action recognition models, such as TPN or VideoST to SlowFast. Also, it demonstrates robust knowledge transfer ability regardless of teacher architecture (TPN or VideoST). Compared to the concurrent method DCSF \cite{dai-aaai2025-channelkd}, ours improves the Top-1 accuracy by 1.11\% in the TPN$\rightarrow$SlowFast task. When compared to classical methods such as KD \cite{geoffrey-arxiv2015-kd} and DKD \cite{zhao-cvpr2022-dkd}, the performance gains are more larger, \ie, 3.10\% and 2.09\%, respectively.

\begin{figure}[!t]
	\centering
		\includegraphics[width=0.95\linewidth]{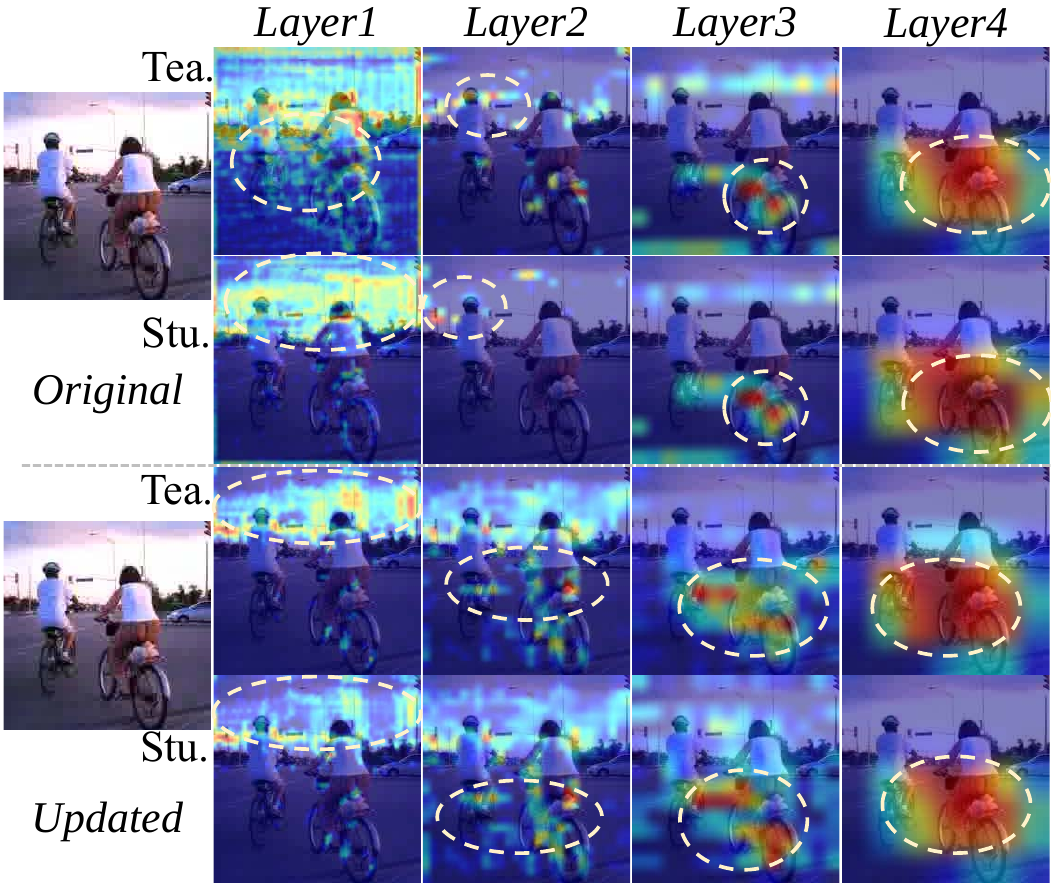}
		\includegraphics[width=0.95\linewidth]{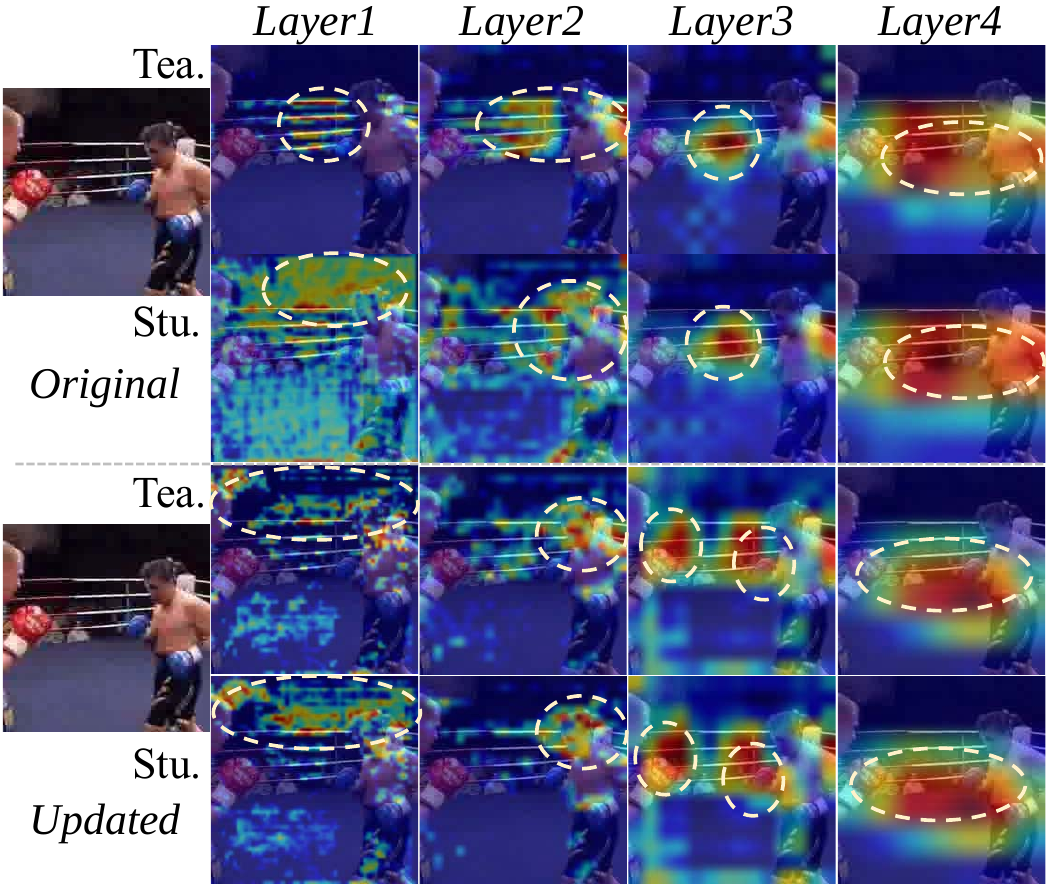}
		\caption{Visualization of feature Class Activation Maps on UCF101 \cite{soomro-arxiv2012-ucf101}. Top: \textit{biking}, bottom: \textit{punch}. }
		\label{fig:vis_ucf}
\end{figure}	

\begin{figure}[!t]
		\centering
		\includegraphics[width=0.95\linewidth]{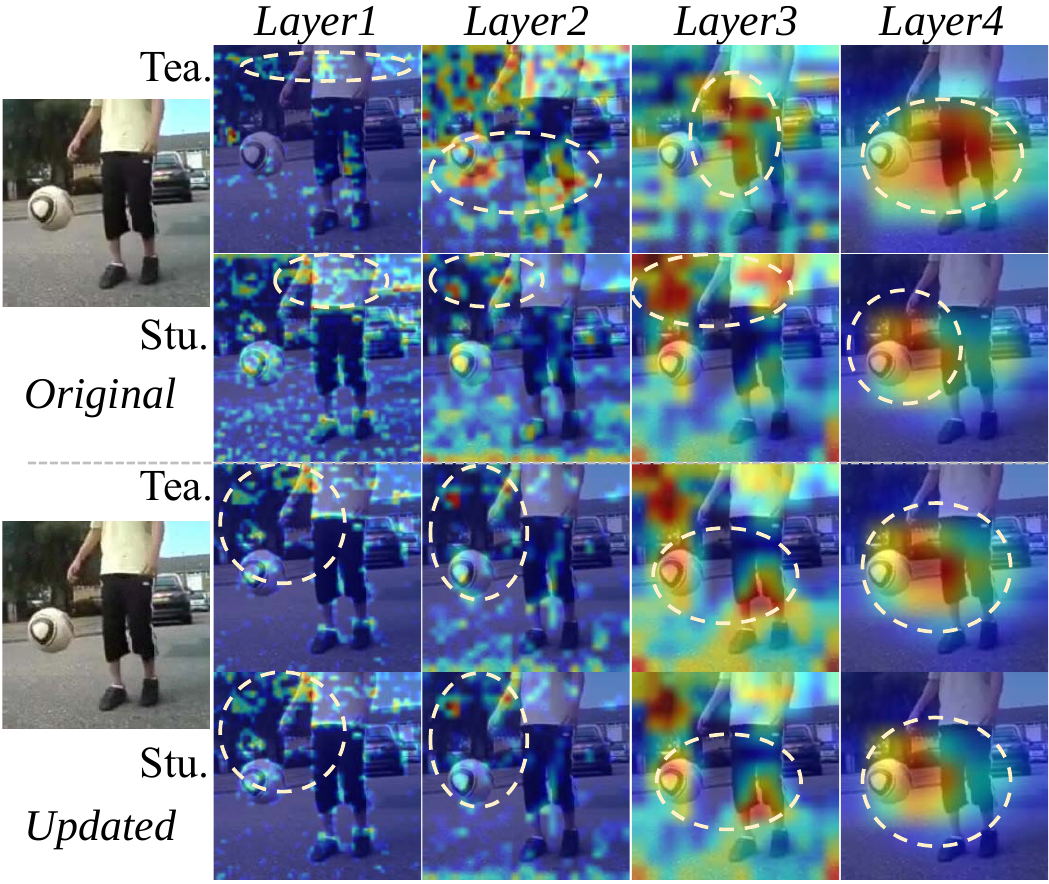}
		\includegraphics[width=0.95\linewidth]{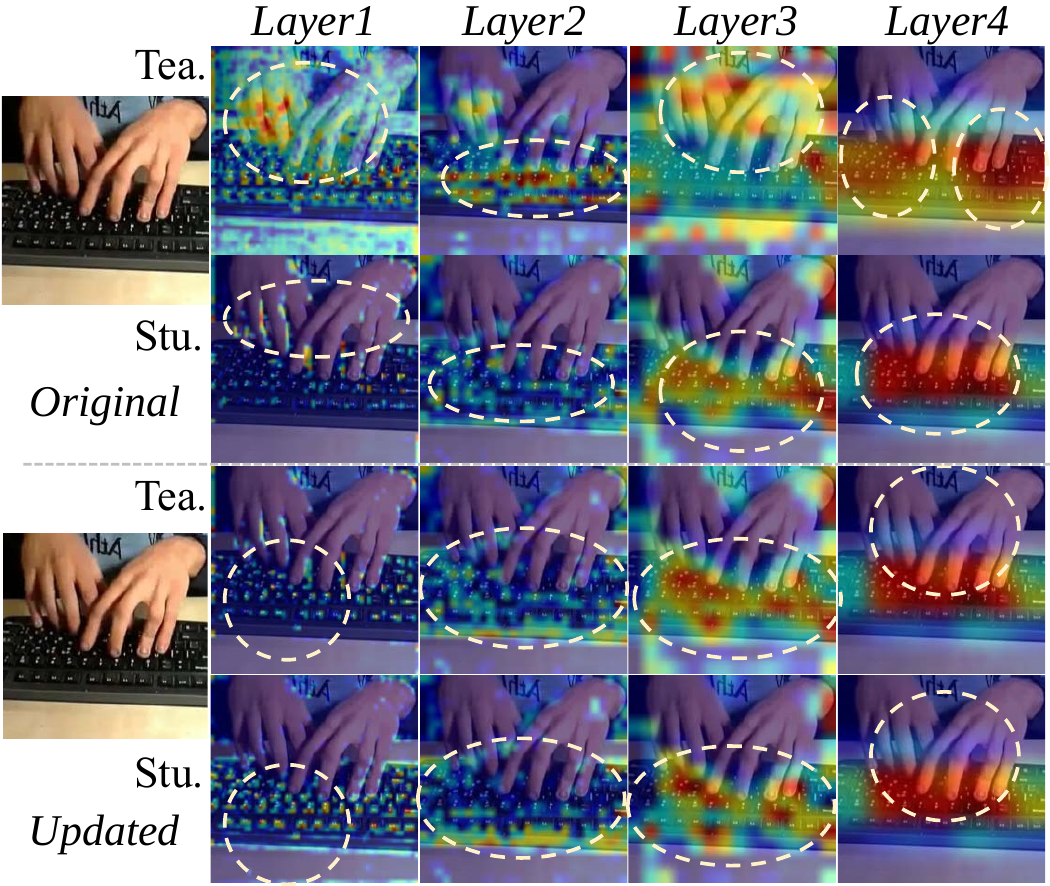}
		\caption{Visualization of feature Class Activation Maps on Kinetics-400 \cite{kay-arXiv2017-kinetics}. Top: \textit{juggling soccer ball}, bottom: \textit{using computer}. }
		\label{fig:vis_kinetics}

\end{figure}

\subsection{Qualitative Results}
To intuitively show feature map differences between the original and updated samples, we randomly select several videos from UCF101 \cite{soomro-arxiv2012-ucf101} and Kinetics-400 \cite{kay-arXiv2017-kinetics}, and show their frame Class Activation Maps (CAM) of intermediate features at \textit{Epoch}~50 across four layers in Fig.~\ref{fig:vis_ucf} and Fig.~\ref{fig:vis_kinetics}, respectively. The model backbone is SlowFast~\cite{feichtenhofer-iccv2019-slowfast}, and experimental setup keeps the same as training. 

As depicted in the figures, the feature maps are better aligned by feeding updated samples to the model for both videos, because the overlapping areas of CAMs at the bottom are much larger than that at the top. This demonstrates that updating samples with guidance of gradients contributes to more satisfying feature-level distillation. In addition, we found that low-level layers tend to capture the backgrounds while high-level layers discover motion semantics such as human-bicycle or human-ball interactions.

\section{Conclusion}
\label{sec:conclu}
This work presents a novel perspective on model compression for action recognition by introducing a feature-based KD framework grounded in an adaptive sample generation manner. Our approach dynamically generates samples that better facilitate the feature alignment, guided by sample gradients. A key point is the projection of video data into the frequency domain, where a Gaussian mask with learnable parameters adaptively preserves motion features while filtering out irrelevant noise. This benefits the computation of optimal gradients for sample updates, which are used to train student using a feature loss weighted by teacher-student centroid frequency differences. Extensive experiments on multiple video and image benchmarks demonstrate the consistent superiority of our method over existing alternatives.


\bibliographystyle{IEEEtran}

\newpage

\appendix

\noindent \textbf{This appendix} provides details about \textit{the overall pipeline, datasets, experimental settings, more ablation studies, discussion on class-imbalance issue, training efficiency analysis, model convergence, and visualization results}. 

\subsection{Overall Pipeline}
Our ASCD framework consists of two main components, \ie, the Adaptive Sample Generation (ASG) module and the Channel-wise Dynamic Distillation (CDD) module, which run alternatively via two stages to achieve the optimal feature alignment. At stage one, both teacher and student are frozen, while updating samples guided by sample gradients governed by learnable frequency and bandwidth in the frequency domain. At stage two, updated samples are fed into frozen teacher and learnable student, while channel-wise feature alignment is dynamically guided by channel centroid frequencies of stage one. 

Specifically, the features of each layer are projected from the spatial domain to the frequency domain along the channel dimension, resulting in channel-wise frequency features. These features enable modeling long-range spatiotemporal dependencies and identifying the redundancy. Meanwhile, the Gaussian mask with learnable centroid frequencies and bandwidths is applied to them for isolating certain-frequency components, such as low-frequency components hold semantic structure while high-frequency ones model fine-grained details (\eg, edges). The masked frequency features are then projected back to the spatial domain for calculating the feature loss, weighted by the reciprocal of centroid frequency difference. We optimize this feature loss by taking the derivatives of varying sample, centroid frequency, and bandwith, leading to the sample gradients $\mathbf{g}$. Note that these gradients are updated in a class-wise manner, \ie, all samples in the same class are used to update and share the gradient. Finally, the samples are updated with corresponding gradients.  

After that, updated samples are fed into teacher and student to obtain their intermediate features and output logits. For feature loss, it is weighted by the channel-wise centroid frequency difference. For logit loss, we evaluate the similarity of class sample gradients, \ie, an inter-class semantic distribution, which guides the logits alignment. To reduce distillation costs, the sample update is trigged in line with a cumulative poly-like strategy \cite{wu-tist2023-poly} periodically.

\subsection{Datasets and Evaluation Metrics}   
We examine the performance of our method on three video benchmarks, \ie, UCF101 \cite{soomro-arxiv2012-ucf101}, Kinetics-400 \cite{kay-arXiv2017-kinetics}, and Sth-Sth-v2 \cite{goyal-iccv2017-sthsth}, and two image benchmarks, \ie, CIFAR-100 \cite{wah-2011-cifar} and ImageNet \cite{krizhevsky-nips2012-imagenet}. Dataset statistics are summarized in Table~\ref{tbl:dataset}. Except for Kinetics-400 using the validation set, the remaining ones use the test set to obtain the results.

\textbf{UCF101}\footnote{https://www.crcv.ucf.edu/data/UCF101.php} dataset~\cite{soomro-arxiv2012-ucf101} consists of action videos collected from YouTube, covering 101 different categories with a total of 13,320 video clips and approximately 27 hours of video duration. It is divided into 25 groups by category, each containing 4 to 7 videos. Each video has a resolution of 320$\times$240 and a frame rate of 25 fps. We follow the official split using Trainlist01 and Testlist01, and all input frames are resized to 224$\times$224 for training.

\textbf{Kinetics-400}\footnote{https://deepmind.com/research/open-source/kinetics} dataset~\cite{kay-arXiv2017-kinetics}, originally released by DeepMind, contains video clips collected from YouTube, covering 400 different action categories, each with at least 400 video samples. Each clip lasts about 10 seconds. The dataset consists of 234,619 training samples and 19,761 validation samples. All input frames are resized to 256$\times$256 for training.

\textbf{Something-Something V2}\footnote{https://www.qualcomm.com/developer/software/something-something-v-2-dataset}~\cite{goyal-iccv2017-sthsth} is a large-scale dataset with 220,847 labeled video clips, depicting humans performing predefined actions using everyday objects. Among them, 168,913 are used for training, 24,777 for validation, and 27,157 for testing. The dataset includes 174 action categories, with video durations ranging from 2 to 6 seconds and an average length of 4.03 seconds. Each category contains approximately 620 videos. Most video frames have a resolution of 240$\times$320 and are uniformly resized to 224$\times$224 for training.

\textbf{CIFAR-100}\footnote{http://www.cs.toronto.edu/~kriz/cifar.html}~\cite{wah-2011-cifar} consists of 100 classes, each containing 600 color images of size 32$\times$32. The dataset is organized into 20 coarse categories, each containing 5 fine-grained classes. It is split into a training set with 50,000 images and a test set with 10,000 images. Each class has 500 training and 100 testing images. 

\textbf{ImageNet}\footnote{https://image-net.org/index.php}~\cite{krizhevsky-nips2012-imagenet} is a subset of the ILSVRC-2012 challenge, containing approximately 1.2 million training images, 50,000 validation images, and 100,000 test images across 1,000 categories. Each category contains around 1,200 images. While image resolutions vary, all inputs are resized to 224$\times$224 before training.

\textbf{Evaluation Metrics}.  Following previous works \cite{zong-iclr2023-kd,li-aaai2023-ctkd,xie-cvpr2023-dynamickd}, we adopt the commonly used Top-1 accuracy and Top-5 accuracy as the evaluation metrics. Also we report the average elapsed time per epoch to show the training efficiency. Top-1 or Top-5 accuracy evaluates those samples whose ground-truth class takes up the top or one of the five leading positions of the candidate class set. 
\begin{table}[!t]
	\centering
	\caption{Statistics of dataset.}
	\label{tbl:dataset}
	\scalebox{0.95}{
		\setlength{\tabcolsep}{1.2mm}{
			\begin{tabular}{l c r  r  rc}
				\toprule[0.75pt]
				Dataset & Class  & Training& Validation & Test  &Resolution\\ 
				\midrule[0.5pt]
				UCF101\cite{soomro-arxiv2012-ucf101}      & 101   & 9,537    & -       & 3,783 &320$\times$240 \\
				Kinetics-400\cite{kay-arXiv2017-kinetics} & 400   & 234,619  & 19,761  & -     &256$\times$256 \\
				Sth-Sth-v2\cite{goyal-iccv2017-sthsth}    & 174   & 168,913  & 24,777  & 27,157 &240$\times$320\\
				CIFAR-100\cite{wah-2011-cifar}            & 100   & 50,000   & -       & 10,000&32$\times$32 \\
				ImageNet\cite{krizhevsky-nips2012-imagenet} & 1,000 & 1,200,000  & 50,000  & 100,000 &224$\times$224\\
				\toprule[0.75pt]
			\end{tabular}
	}}
\end{table}

\begin{table}[!t]
	\centering
	\caption{Batch size and training epoch.}
	\label{tbl:batch_epoch_setting}
	\scalebox{0.95}{
		\setlength{\tabcolsep}{0.6mm}{
			\begin{tabular}{lccc  cccc}
				\toprule[0.75pt]
				\multirow{3}{*}{Model} & \multicolumn{3}{c}{Batch Size} && \multicolumn{3}{c}{$N_{max}$} \\ 
				\cmidrule[0.5pt]{2-4} \cmidrule[0.5pt]{6-8} 
				& UCF101 & Kinetics & Sth-Sth-v2 && UCF101 & Kinetics & Sth-Sth-v2 \\
				\midrule[0.5pt]
				SlowFast \cite{feichtenhofer-iccv2019-slowfast} & 32 & 48 & 48 && 50 & 200 & 200 \\
				TPN \cite{yang-cvpr2020-temporal} & 4 & - & - && 50 & - & - \\
				VideoST \cite{liu-cvpr2022-video} & 4 & 8 & 8 && 100 & 200 & 200 \\
				\toprule[0.75pt]
			\end{tabular} 
		}
	}
\end{table}

\subsection{Experimental Settings}
\textbf{Training}. For UCF101 \cite{soomro-arxiv2012-ucf101}, the model parameters for action recognition are initialized with the model pre-trained on Kinetics-400 \cite{kay-arXiv2017-kinetics}, there are 32 sampled frames in a clip. For Kinetics-400 and Sth-Sth-v2, there are 16 sampled frames in a clip. The Stochastic Gradient Descent optimizer is used with a momentum of 0.9. The initial learning rate for videos is 1e-2, with a weight decay factor of 1e-4 after each epoch. For Kinetics-400 and Sth-Sth-v2, following \cite{liu-cvpr2022-video}, both the tiny version (SwinT) and the small version (SwinS) of Video Swin Transformer (Video ST) \cite{liu-cvpr2022-video} are pre-trained on ImageNet, while models are randomly initialized. For CIFAR-100 and ImageNet \cite{krizhevsky-nips2012-imagenet}, following \cite{pham-wacv2024-frequency}, it has an initial learning rate of 0.1, a decay factor of 0.1 applied at epochs 150, 180, and 210, the batch size of 256; the maximum epoch is 240 and 200, respectively. For clarity, the batch size and training epochs of three video models on three video benchmarks are shown in Table~\ref{tbl:batch_epoch_setting}.

For our method, the hyper-parameters of KD loss terms are: $\alpha$=100, $\beta$=0.9, $\gamma$=0.5; $\lambda$ in sample generation is 0.9 for video and 0.1 for image; $\eta$ in gain function is 0.7 for video and 0.3 for image. The number $L$ of intermediate layers is 4 (\eg, 4 stages for ResNet). For other methods, we run the publicly available codes from original papers to report results. Since codes of DIST+ \cite{huang-tpami2025-dist} and DualKD \cite{wang-tip2024-dkd} are unavailable, we try the best to implement them by ourself.

\noindent\textbf{Inference}. Given a test sample, it is normalized before passing into student model to output the class probabilities.

\noindent\textbf{Video/image models}. To verify the generalization ability of our method on both video and image samples, we examine KD methods on three action recognition models, including SlowFast \cite{feichtenhofer-iccv2019-slowfast}, TPN (Temporal Pyramid Network) \cite{yang-cvpr2020-temporal}, and VideoST (Video SwinTransformer) \cite{liu-cvpr2022-video}, as well as three image models (settings follow \cite{pham-wacv2024-frequency}), including ResNet (Residual Neural Network) \cite{he-cvpr2016-residual}, WRN (Wide ResNet) \cite{zagoruyko-bmvc2017-residual}, and MobileNetv2 \cite{sandler-cvpr2018-mobilenetv2}. Following \cite{li-mm2025-sakd}, for SlowFast, we sample 16 or 8 or 4 frames in a clip when the step is set to 8 or 16, termed SF16x8 or SF8x8 or SF4x16; for TPN teacher or student, we sample 32 or 8 frames temporally and scale up spatially by a factor of 2 or 8, termed TPN-f32s2 or TPN-f8s8. For VideoST, Transformer layer at Stage~3 of SwinS/SwinT is 18/6 as teacher/student. For SlowFast and TPN, teacher and student adopt ResNet101 and ResNet50 as the backbone respectively. The parameters of action recognition or image classification models are kept the same as original papers. Their computational costs are reported in Table~\ref{tbl:backbone_flops_size}.

\begin{table}[!t]
	\centering
	\small 
	\caption{Computational costs of basic models.}
	\scalebox{0.9}{
		\setlength{\tabcolsep}{0.3mm}{
			\begin{tabular}{llccrrr}
				\toprule
				\multirow{2}{*}{Basic Model} & \multirow{2}{*}{Backbone} & \multirow{2}{*}{Teac.} & \multirow{2}{*}{Stud.}& \multicolumn{3}{c}{Model Size}  \\ 
				\cmidrule(lr{0.5pt}){5-7}
				& &&  & Params(M)$\downarrow $ & FLOPs(G)$\downarrow$ & FPS$\uparrow$ \\
				\midrule
				\multirow{2}{*}{TPN \cite{yang-cvpr2020-temporal}}  & TPN-f32s2  &\checkmark& &  99.71 &375.09  & 298  \\
				& TPN-f8s8  &&\checkmark  & 71.80  &202.05 & 516  \\
				\midrule[0.5pt]
				\multirow{3}{*}{SlowFast\cite{feichtenhofer-iccv2019-slowfast} }  & SF16$\times$8&\checkmark&   &62.14  &97.19 & 953\\
				& SF8$\times$8& \checkmark&   &34.57 &50.86  &1482  \\
				& SF4$\times$16 && \checkmark &33.79  &28.01  &2125  \\
				\midrule[0.5pt]
				\multirow{2}{*}{VideoST\cite{liu-cvpr2022-video}}&SwinS &\checkmark   &  &49.10  &138.14 &361  \\
				& SwinT & &\checkmark & 27.81 &70.91 & 597  \\
				\midrule[0.5pt]
				\multirow{3}{*}{WideResNet\cite{zagoruyko-bmvc2017-residual} }  & WRN40-2  &\checkmark   & &2.26  &0.33  & 267 \\
				& WRN40-1 && \checkmark   &0.57  & 0.08 &202 \\
				& WRN16-2 && \checkmark   &0.70  &0.10  & 515\\
				\midrule[0.5pt]
				\multirow{6}{*}{	ResNet\cite{he-cvpr2016-residual} }  &RN110&\checkmark   & &1.74  &0.26 &104 \\
				&RN56&\checkmark               &      &0.86  &0.13  & 204 \\
				& RN32$\times$4&\checkmark     &      &7.43  &1.09  &324  \\
				& RN32&& \checkmark            &0.47  &0.07  &249  \\
				& RN20&& \checkmark            &0.28  & 0.04 &384  \\
				& RN8$\times$4 && \checkmark   &1.23  &0.18  & 722 \\
				&RN50&\checkmark    &           &23.56  &4.10 &75  \\
				&RN34&\checkmark    &           &21.81  &3.60 &86  \\
				&RN18&    &  \checkmark         &11.79  &1.80 &176 \\
				\midrule[0.5pt]
				\multirow{1}{*}{	MobileNet\cite{sandler-cvpr2018-mobilenetv2} }  &	MobileNetv2&   &\checkmark &1.7  &0.18 &328\\
				\bottomrule
			\end{tabular}
		}
	}
	\label{tbl:backbone_flops_size}
\end{table}

\subsection{Compared Methods}
We compare our method with two groups of SOTA KD methods in the following:

1) \textit{logit-based} methods: DKD (Decoupled KD) \cite{zhao-cvpr2022-dkd}, CTKD (Curriculum Temperature KD) \cite{li-aaai2023-ctkd}, CrossKD (Cross-head KD) \cite{wang-cvpr2024-crosskd}, and SAKD (Sample-level Adaptive KD) \cite{li-mm2025-sakd}.

2) \textit{feature-based} methods: DualKD \cite{wang-tip2024-dkd}, CKD (Channel-wise KD) \cite{shu-iccv2021-channelkd}, GKD (Generative model based KD) \cite{wang-aaai2024-gkd}, DCSF (Dynamic Channel-wise Spatial Feature KD) DCSF \cite{dai-aaai2025-channelkd}, and DIST+ \cite{huang-tpami2025-dist}.  

\subsection{More Ablation Studies}
This section provides more ablations of $\lambda$ and $\eta$ in the Adaptive Sample Generation (ASG) module as well as $\alpha$ and $\gamma$ in the Channel-wise Dynamic Distillation (CDD) module on the remaining two datasets, \ie, Sth-Sth-v2 \cite{goyal-iccv2017-sthsth} and CIFAR-100 \cite{wah-2011-cifar}. Besides, we provide ablations on gain schedule for sample update, inter-class similarity calculation, feature alignment forms at stage one, as well as cross-modal knowledge distillation. Note that experimental settings keep the same as in training, unless specified. 

\begin{table}[!h]
	\centering
	\caption{Ablations of $\lambda$ on Sth-Sth-v2 \cite{goyal-iccv2017-sthsth} and CIFAR-100 \cite{wah-2011-cifar}.}
	\small
	\scalebox{0.8}{
		\setlength{\tabcolsep}{0.63mm}{
			\begin{tabular}{c ccccccc cccccc}
				\toprule[0.75pt]
				\multirow{3}{*}{$\lambda$} && \multicolumn{5}{c}{Sth-Sth-v2\cite{goyal-iccv2017-sthsth}} &&\multicolumn{5}{c}{CIFAR-100\cite{wah-2011-cifar}} &\\
				\cmidrule(lr{0.5pt}){2-7} \cmidrule(lr{0.5pt}){8-13}
				&&\multicolumn{2}{c}{Top1$\uparrow$} & &\multicolumn{2}{c}{Top5$\uparrow$} &&\multicolumn{2}{c}{Top1$\uparrow$} & &\multicolumn{2}{c}{Top5$\uparrow$} & \\
				\cmidrule(lr{0.5pt}){2-4} \cmidrule(lr{0.5pt}){5-7} \cmidrule(lr{0.5pt}){8-10} \cmidrule(lr{0.5pt}){11-13}
				&& SlowFast & VideoST && SlowFast & VideoST && ResNet	&WRN & & ResNet	&WRN&\\
				\midrule[0.5pt]
				0.0 &&46.89 & 56.98&&75.95 &86.44 && 76.82 & \underline{76.53} & & 93.81 & \underline{94.91} &\\
				0.1 &&46.93 & 57.02&&75.99 &86.52 && \bf77.95 &\bf76.73 & & \bf95.02 & \bf95.03 & \\
				0.3 &&47.08 & 57.15&&76.00 &86.68 && \underline{77.62} & 76.15 & & \underline{94.42} & 94.10 &\\
				0.5 &&47.18 & 57.24&&75.41 &86.79 && 77.02 & 75.80 & & 93.92 & 93.75 &\\
				0.7 &&\underline{47.25} & \underline{57.28}&&\bf76.42 &\underline{86.83} && 76.82 & 75.48 & & 93.60 & 93.43 &\\
				0.9 && \textbf{47.28} & \textbf{57.31}&& \underline{76.03} & \textbf{86.87} && 76.25 & 75.21 & & 93.23 & 93.16 &\\
				1.0 &&47.10 & 57.20&&76.01 &86.75 && 76.13 & 74.99 & & 92.04 & 92.94 &\\
				\toprule[0.75pt]
			\end{tabular}
		}
	}
	\label{tbl:lambda-sth-cifar}
\end{table}

\begin{table}[!t]
	\centering
	\caption{Ablations of $\eta$ on Sth-Sth-v2 \cite{goyal-iccv2017-sthsth} and CIFAR-100 \cite{wah-2011-cifar}.}
	\small
	\scalebox{0.8}{
		\setlength{\tabcolsep}{0.63mm}{
			\begin{tabular}{c ccccccc cccccc}
				\toprule[0.75pt]
				\multirow{3}{*}{$\eta $} && \multicolumn{5}{c}{Sth-Sth-v2\cite{goyal-iccv2017-sthsth}} &&\multicolumn{5}{c}{CIFAR-100\cite{wah-2011-cifar}} &\\
				\cmidrule(lr{0.5pt}){2-7} \cmidrule(lr{0.5pt}){8-13}
				&&\multicolumn{2}{c}{Top1$\uparrow$} & &\multicolumn{2}{c}{Top5$\uparrow$} &&\multicolumn{2}{c}{Top1$\uparrow$} & &\multicolumn{2}{c}{Top5$\uparrow$} & \\
				\cmidrule(lr{0.5pt}){2-4} \cmidrule(lr{0.5pt}){5-7} \cmidrule(lr{0.5pt}){8-10} \cmidrule(lr{0.5pt}){11-13}
				&& SlowFast & VideoST && SlowFast & VideoST && ResNet &WRN & & ResNet &WRN&\\
				\midrule[0.5pt]
				0.0 &&46.55&56.45 & & 75.34& 85.13 &&76.43 &75.56 & &93.71 &94.68 &\\
				0.1 &&46.83&56.51&&75.59&86.29&&76.89 &76.02 & &94.15 &94.35 &\\
				0.3 &&47.11&56.83&&75.72  & 86.54&& \bf77.95 & \bf76.73 & & \bf95.02 & \bf{94.52} & \\	
				0.5 &&\underline{47.19}&\underline{57.01}&& \underline{75.95}& \underline{86.83}& & \underline{77.68} & \underline{76.45} & & \underline{94.87} & \underline{94.30} & \\	
				0.7 &&\textbf{47.28} & \textbf{57.31}&& \bf{76.03} & \textbf{86.87}&  &77.32 & 76.18 & & 94.65 & 94.25 & \\	
				0.9 &&46.92  &56.72& &75.45&86.02&&77.05 &75.92 & &94.35 &94.10 & \\	
				1.0 &&46.73&56.61& & 75.28& 85.89 &&76.78 &75.74 & &94.20 &93.95 & \\	
				\toprule[0.75pt]
			\end{tabular}
		}
	}
	\label{tbl:eta-sth-cifar}
\end{table}

\textbf{$\lambda$ and $\eta$ in ASG}. To investigate the influences of the sample update hyper-parameter $\lambda$ and the gain magnitude hyper-parameter $\eta$, we vary them from 0 to 1.0 with seven bins and their results are respectively shown in Table~\ref{tbl:lambda-sth-cifar} and Table~\ref{tbl:eta-sth-cifar}. From the records, their optimal values are dataset-dependent, \eg, the video dataset (Sth-Sth-v2) prefers higher values, and the image dataset (CIFAR-100) prefers lower to medium values. They share the similar tendency that performance generally improves as the parameters increase from 0, peaks at a specific range, and then declines. For example, on Sth-Sth-v2, performance improves as $\eta$ increases, reaching an optimum at $\eta = 0.7$; on CIFAR-100, performance peaks at a much lower value of $\eta = 0.3$ and then gradually declines as $\eta$ increases further. 

\begin{table}[!t]
	\centering
	\caption{Ablations of $\alpha$ on Sth-Sth-v2 \cite{goyal-iccv2017-sthsth} and CIFAR-100 \cite{wah-2011-cifar}.}
	\small
	\scalebox{0.8}{
		\setlength{\tabcolsep}{0.63mm}{
			\begin{tabular}{c ccccccc cccccc}
				\toprule[0.75pt]
				\multirow{3}{*}{$\alpha$} && \multicolumn{5}{c}{Sth-Sth-v2\cite{goyal-iccv2017-sthsth}} &&\multicolumn{5}{c}{CIFAR-100\cite{wah-2011-cifar}} &\\
				\cmidrule(lr{0.5pt}){2-7} \cmidrule(lr{0.5pt}){8-13}
				&&\multicolumn{2}{c}{Top1$\uparrow$} & &\multicolumn{2}{c}{Top5$\uparrow$} &&\multicolumn{2}{c}{Top1$\uparrow$} & &\multicolumn{2}{c}{Top5$\uparrow$} & \\
				\cmidrule(lr{0.5pt}){2-4} \cmidrule(lr{0.5pt}){5-7} \cmidrule(lr{0.5pt}){8-10} \cmidrule(lr{0.5pt}){11-13}
				&& SlowFast & VideoST && SlowFast & VideoST && ResNet & WRN && ResNet & WRN &\\
				\midrule[0.5pt]
				10 &&46.37&56.45 & & 75.67& 85.68 &&76.66&75.58&&94.55&94.07&\\
				30 &&46.43&56.61&& 75.71 &86.72 &&77.31&76.26&&94.69&94.23&\\
				75 &&\underline{46.73}&\underline{56.93}&&\underline{75.91}  &\underline{86.85}  &&\underline{77.72}&\underline{76.68}&&\underline{94.89}&94.76&\\
				100 &&\textbf{47.28} & \textbf{57.31}&& \bf{76.03} & \textbf{86.87}& &\bf77.95&\bf76.73&&\bf95.02&\underline{94.52}&\\
				150 &&46.82&56.83& &75.93 &86.74 &&77.23&76.48&&94.83&\bf94.72&\\
				300 &&46.65  &56.68& &75.74&86.52&&76.54&75.92&&94.37&94.21&\\
				500 &&46.39&56.37& &75.58 &85.96  &&75.48&75.62&&94.19&93.98&\\
				\toprule[0.75pt]
			\end{tabular}
		}
	}
	\label{tbl:alpha-sth-cifar}
\end{table}

\begin{table}[!t]
	\centering
	\caption{Ablations of $\gamma$ on Sth-Sth-v2 \cite{goyal-iccv2017-sthsth} and CIFAR-100 \cite{wah-2011-cifar}.}
	\small
	\scalebox{0.8}{
		\setlength{\tabcolsep}{0.63mm}{
			\begin{tabular}{c ccccccc cccccc}
				\toprule[0.75pt]
				\multirow{3}{*}{$\gamma$} && \multicolumn{5}{c}{Sth-Sth-v2\cite{goyal-iccv2017-sthsth}} &&\multicolumn{5}{c}{CIFAR-100\cite{wah-2011-cifar}} &\\
				\cmidrule(lr{0.5pt}){2-7} \cmidrule(lr{0.5pt}){8-13}
				&&\multicolumn{2}{c}{Top1$\uparrow$} & &\multicolumn{2}{c}{Top5$\uparrow$} &&\multicolumn{2}{c}{Top1$\uparrow$} & &\multicolumn{2}{c}{Top5$\uparrow$} & \\
				\cmidrule(lr{0.5pt}){2-4} \cmidrule(lr{0.5pt}){5-7} \cmidrule(lr{0.5pt}){8-10} \cmidrule(lr{0.5pt}){11-13}
				&& SlowFast & VideoST && SlowFast & VideoST && ResNet	&WRN & & ResNet	&WRN&\\
				\midrule[0.5pt]
				0.0 &&46.55&56.45 & & 75.84& 85.93 &&76.54&75.84 & &94.45 &93.82 &\\
				0.1 &&46.72&56.72&& 75.62 &86.21 &&77.09&76.15 & &94.62 &94.23 &\\
				0.3 &&\underline{47.23}&\underline{56.86}&&\underline{75.93}  &86.35  &&77.47& \underline{76.62}& & \underline{94.86} &94.49 &\\
				0.5 &&\textbf{47.28} & \textbf{57.31}&& \bf{76.03} &   \textbf{86.87} &&\bf77.95&  \bf76.73&	&\bf95.02&\underline{94.52} &\\
				0.7 &&46.93&56.72& &75.85 &\underline{86.37}  &&\underline{77.68}&76.48 & &94.73 &\bf94.83 &\\
				0.9 &&46.61  &56.48& &75.63&86.02&&77.05&75.89 & &94.51 &94.32 &\\
				1.0 &&46.24&56.24& &75.42 &85.47  &&76.73&75.58 & &94.58&94.02 &\\
				\toprule[0.75pt]
			\end{tabular}
		}
	}
	\label{tbl:gamma-sth-cifar}
\end{table}

\textbf{$\alpha$ and $\gamma$ in CDD}. To investigate the influences of the regularization parameter $\alpha$ for the feature alignment loss $\mathcal{L}_{kd}^{fea2}$ and $\gamma$ for the class similarity guided KL loss, we vary $\alpha$ from 10 to 500 and $\gamma$ from 0 to 1, whose results are respectively recorded in Table~\ref{tbl:alpha-sth-cifar} and Table~\ref{tbl:gamma-sth-cifar}. From the records, we observe the optimal values for $\alpha$ (100) and $\gamma$ (0.5) are identical for both the video and image datasets. This suggests that the CDD component is robust and its hyper-parameters are easily to be transferred across different data domains. Moreover, their values cannot be too small or too large in their ranges. For $\alpha$ on video dataset, performance improves steadily as $\alpha$ increases from 10 to 100, achieving peak results across all metrics (Top-1 and Top-5 for both SlowFast and VideoST), \eg, VideoST Top-1 accuracy peaks at 57.31\%. For $\gamma$ on video dataset, performance improves as $\gamma$ increases from 0, reaching a clear peak at $\gamma = 0.5$, which is consistent for both Top-1 and Top-5 accuracy on both video architectures, \ie, SlowFast and VideoST; on image dataset, the trend is consistent with the video, when the best results are achieved at $\gamma = 0.5$ for both ResNet and WRN models.

\begin{table}[!t]
	\centering
	\caption{Ablations of gain schedule for sample update on Sth-Sth-v2 \cite{goyal-iccv2017-sthsth} and CIFAR-100 \cite{wah-2011-cifar}.}
	\small
	\scalebox{0.8}{
		\setlength{\tabcolsep}{0.5mm}{
			\begin{tabular}{l ccccccc cccccc}
				\toprule[0.75pt]
				\multirow{3}{*}{Method} &&  \multicolumn{5}{c}{Sth-Sth-v2\cite{goyal-iccv2017-sthsth}} &&\multicolumn{5}{c}{CIFAR-100\cite{wah-2011-cifar}} &\\
				\cmidrule(lr{0.5pt}){2-7}  \cmidrule(lr{0.5pt}){8-13}
				&&\multicolumn{2}{c}{Top1$\uparrow$} & &\multicolumn{2}{c}{Top5$\uparrow$} &&\multicolumn{2}{c}{Top1$\uparrow$} & &\multicolumn{2}{c}{Top5$\uparrow$} &  \\ 
				\cmidrule(lr{0.5pt}){2-4}  \cmidrule(lr{0.5pt}){5-7} \cmidrule(lr{0.5pt}){8-10}  \cmidrule(lr{0.5pt}){11-13}
				&& SlowFast	& VideoST && SlowFast	& VideoST && ResNet	&WRN & & ResNet	&WRN&\\ 
				\midrule[0.5pt]
				Linear	&&  \underline{46.87} &\underline{56.94}&	&\bf76.09	&\underline{86.28}		&	&\underline{77.83}	&\underline{76.64} &  &\underline{94.83}	&	\bf94.93		& \\	
				Cosine	&& 46.72 &56.82&&	75.91&	86.04&			&76.63	&75.83&  &94.38	&	94.21		& \\	
				Poly	&&\textbf{47.28} & \textbf{57.31}&& \underline{76.03} & \textbf{86.87}&&\bf77.95	&  \bf76.73&	&\bf95.02	&\underline{94.52}			 		& \\
				\toprule[0.75pt]
			\end{tabular}
		}
	}
	\label{tbl:gain-sth-cifar}
\end{table}

\textbf{Gain Schedule in ASG}. To examine the different gain schedules, we adopt three forms including Linear, Cosine, and Poly, whose results are shown in Table~\ref{tbl:gain-sth-cifar}. Across both video and image datasets (Sth-Sth-v2, and CIFAR-100) and all model architectures, the Poly schedule consistently delivers the best or very competitive performance. The Linear schedule is almost always the second-best option, often very close to the top results and sometimes slightly outperforming Poly on a single metric (\eg, SlowFast Top-5 on Sth-Sth-v2). 

\begin{table}[!t]
	\centering
	\caption{Ablations of inter-class similarity calculation on Sth-Sth-v2 \cite{goyal-iccv2017-sthsth} and CIFAR-100 \cite{wah-2011-cifar}.}
	\small
	\scalebox{0.7}{
		\setlength{\tabcolsep}{0.63mm}{
			\begin{tabular}{l ccccccc cccccc}
				\toprule[0.75pt]
				\multirow{3}{*}{Similarity} &&  \multicolumn{5}{c}{Sth-Sth-v2\cite{goyal-iccv2017-sthsth}} &&\multicolumn{5}{c}{CIFAR-100\cite{wah-2011-cifar}} &\\
				\cmidrule(lr{0.5pt}){2-7}  \cmidrule(lr{0.5pt}){8-13}
				&&\multicolumn{2}{c}{Top1$\uparrow$} & &\multicolumn{2}{c}{Top5$\uparrow$} &&\multicolumn{2}{c}{Top1$\uparrow$} & &\multicolumn{2}{c}{Top5$\uparrow$} &  \\ 
				\cmidrule(lr{0.5pt}){2-4}  \cmidrule(lr{0.5pt}){5-7} \cmidrule(lr{0.5pt}){8-10}  \cmidrule(lr{0.5pt}){11-13}
				&& SlowFast	& VideoST && SlowFast	& VideoST && ResNet	&WRN & & ResNet	&WRN&\\ 
				\midrule[0.5pt]
				Mahalanobis 	&& \underline{46.93}&56.83& &\bf76.92&	\underline{86.83}		&	&77.83	& 75.83&  &94.25	&94.45		& \\	
				RBF Kernel  		&& 46.62 &\underline{57.03}&&75.82&86.34			&	&77.43	&74.92 &  &94.92	&94.37		& \\	
				Gaussian kernel	&& 46.72 &56.69&&75.88		&85.48	&	&76.93	&\underline{76.28} &  &94.27	&\underline{94.85}	& \\	
				Cosine &&\textbf{47.28} & \textbf{57.31}&& \underline{76.03} & \textbf{86.87}&&\bf77.95	&  \bf76.73&	&\bf95.02	&\bf{94.52}			 		& \\
				\toprule[0.75pt]
			\end{tabular}
		}
	}
	\label{tbl:clasim-sth-cifar}
\end{table}

\textbf{Inter-class Similarity Calculation}. We examine four different forms of calculating the inter-class similarity in terms of class gradient, which captures the discriminant structure by student. The results are shown in Table~\ref{tbl:clasim-sth-cifar}. From the table, the Cosine similarity scheme consistently outperforms all other methods (Mahalanobis, RBF Kernel, Gaussian Kernel) across different datasets and model architectures. In particular, it delivers state-of-the-art results on challenging datasets, \eg, 57.31\% Top-1 on Sth-Sth-v2 (VideoST) and 77.95\% Top-1 on CIFAR-100 (ResNet). Moreover, the Mahalanobis similarity method shows competitive results on some video cases but inconsistent across datasets.

\textbf{Cross-modal KD}. We examine the knowledge distillation performance under cross-modal settings, \ie, heterogeneous architectures for teacher and student, including Video SwinTransformer \cite{liu-cvpr2022-video} to Video Mamba \cite{li-eccv2024-videomamba}, Transformer from large to small size, and Transformer to CNN. The results on UCF101 are shown in Table~\ref{tbl:ucf_crossmodal}, which involves Video SwinTransformer-Big (Swin-B, Params: 87.28M, FLOPs: 281.61G), Video SwinTransformer-Tiny (Swin-T, 27.81M/70.91G), Video Mamba-M (VM-M, 35.79M/56.51G) , Video Mamba-Tiny (VM-T, 7.15M/11.22G), and SlowFast 4x16 (SF4-16, 33.79M/28.01G) \cite{feichtenhofer-iccv2019-slowfast}. Note that on top row, the former is teacher and the latter is student. From the table, our method demonstrate excellent adaptability for the Mamba architecture. For example, in the same-architecture distillation scenario (VM-M$\rightarrow$VM-T), the student achieves a leap in Top-1 accuracy from 87.20\% to 89.13\%, \ie, a gain of 1.21\% over the state-of-the-art SAKD method. Furthermore, in the cross-model distillation (Swin-B$\rightarrow$VM-T), our method still delivers a 1.25\% Top-1 improvement. This validates the robustness of dynamic sample generation and channel-aware distillation across different modeling architectures.

\begin{table*}[!t]
	\centering
	\caption{Cross-modal KD performance on UCF101 \cite{soomro-arxiv2012-ucf101}.}
	\label{tbl:ucf_crossmodal}
	\scalebox{0.9}{ 
		\setlength{\tabcolsep}{3.5mm}{
			\begin{tabular}{ll cc cc cc cc}
				\toprule[0.75pt]
				\multirow{2}{*}{Method} & \multirow{2}{*}{Venue} &  
				\multicolumn{2}{c}{VM-M/VM-T} &  
				\multicolumn{2}{c}{Swin-B/Swin-T} &
				\multicolumn{2}{c}{Swin-B/VM-T} &
				\multicolumn{2}{c}{Swin-B/SF4-16} \\ 
				\cmidrule(lr){3-4} \cmidrule(lr){5-6} \cmidrule(lr){7-8} \cmidrule(lr){9-10}
				& & Top-1$\uparrow$ & Top-5$\uparrow$ & Top-1$\uparrow$ & Top-5$\uparrow$ & Top-1$\uparrow$ & Top-5$\uparrow$ & Top-1$\uparrow$ & Top-5$\uparrow$ \\
				\midrule[0.5pt]
				Teac.  & - & 92.49 & 98.73 & 89.29 & 98.21 &  89.29 & 98.21 & 89.29 & 98.21 \\
				Stud. & - & 87.20  & 97.35  & 84.40 & 96.17 & 87.20 & 97.35 & 83.13 & 95.72 \\
				\midrule[0.5pt]
				CKD \cite{shu-iccv2021-channelkd} & ICCV'21 & 87.41 & 97.38 & 84.85 & 96.35 & 87.25 & 97.38 & 84.02 & 96.12 \\
				DKD \cite{zhao-cvpr2022-dkd} & CVPR'22 & 87.52 & 97.41 & 85.23 & 96.58 & 87.31 & 97.40 & 84.55 & 96.38 \\
				CTKD \cite{li-aaai2023-ctkd} & AAAI'23 & 87.65 & 97.44 & 85.90 & 96.82 & 87.39 & 97.42 & 85.41 & 96.55 \\
				GKD \cite{wang-aaai2024-gkd} & AAAI'24 & 87.73 & 97.47 & 86.42 & 97.05 & 87.45 & 97.43 & 86.20 & 96.71 \\
				CrossKD \cite{wang-cvpr2024-crosskd} & CVPR'24 & 87.80 & 97.49 & 87.15 & 97.18 & 87.51 & 97.44 & 87.05 & 96.88 \\
				DualKD \cite{wang-tip2024-dkd} & TIP'24 & 87.84 & 97.50 & 87.56 & 97.29 & 87.55 & 97.45 & 87.42 & 97.02 \\
				DCSF \cite{dai-aaai2025-channelkd} & AAAI'25 & 87.88 & \underline{97.51} & 87.72 & 97.35 & 87.59 & 97.45 & 87.71 & 97.11 \\
				SAKD\cite{li-mm2025-sakd} & MM'25 & 87.92  & 97.49  & \underline{87.94}  & \underline{97.42}  & 87.62 & \underline{97.45}  & \underline{87.93}  & \underline{97.19} \\
				DIST+ \cite{huang-tpami2025-dist} & TPAMI'25 & \underline{88.05} & 97.50 & 87.88 & 97.38 & \underline{87.75} & 97.44 & 87.85 & 97.15 \\
				\midrule[0.5pt]
				Ours & - & \bf{89.13} & \bf{98.12} & \bf{89.12} & \bf{98.37} & \bf{88.87} & \bf{97.98} &\bf{88.67} &\bf{97.88}  \\
				Gain & - & \textcolor{red}{+1.08} &\textcolor{red}{+0.61}& \textcolor{red}{+1.18} & \textcolor{red}{+0.95} & \textcolor{red}{+1.12} & \textcolor{red}{+0.53} & \textcolor{red}{+0.74} &\textcolor{red}{+0.69} \\ 
				\bottomrule[0.75pt]
			\end{tabular}
		}
	}
\end{table*}

\textbf{Details of frequency-domain processing}. We probe into the detailed processing submodules in frequency domain, including Freq (Frequency-domain processing in ASG, \textcolor{blue}{see Eqs.~(4)-(7) in the paper}), GBank (Gradient Bank, \textcolor{blue}{see Eq.~(9) in the paper}), SG (Sample Generation, \textcolor{blue}{see Eqs.~(10)-(11) in the paper}), and DyD (Dynamic Distillation, \textcolor{blue}{see Eq.~(12) in the paper}), whose results are shown in Table \ref{tbl:abl_frequency}. Note that the Freq and DyD submodules are coupled together, \ie, it is impossible to do dynamic distillation without frequency-domain pre-processing. From the table, the vanilla model achieves only 84.82\% Top-1 accuracy using SlowFast, our full ASCD framework pushes this performance to 88.42\% by a gain of 3.6\%. In particular, the absence of Freq and DyD submodules leads to a marked decline in performance, underscoring the critical role of frequency-domain feature alignment and channel-wise dynamic weight adjustment in capturing complex video action semantics. Meanwhile, without gradient bank, the performance deteriorates in some degree, which indicates the importance of history sample gradients. Moreover, the SG submodule notably improves accuracy, although it incurs additional computational overhead per sample update. In addition, without dynamic distillation, it worsens the performance by a large margin, which suggests weighting the feature loss in a dynamic way is beneficial for transferring knowledge from teacher to student.  


\begin{table*}[!t]
	\centering
	\caption{Ablations of frequency-domain processing.}
	\label{tbl:abl_frequency}
	\scalebox{1.0}{ 
		\setlength{\tabcolsep}{2.5mm}{ 
			\begin{tabular}{cccc cc cc cc cc cc cc}
				\toprule[0.75pt]
				\multirow{3}{*}{Freq} & \multirow{3}{*}{GBank} & \multirow{3}{*}{SG} & \multirow{3}{*}{DyD} & \multicolumn{4}{c}{UCF101 \cite{soomro-arxiv2012-ucf101}} & \multicolumn{4}{c}{Kinetics-400 \cite{kay-arXiv2017-kinetics}} & \multicolumn{4}{c}{Sth-Sth-v2 \cite{goyal-iccv2017-sthsth}} \\
				\cmidrule(lr){5-8} \cmidrule(lr){9-12} \cmidrule(lr){13-16}
				
				& & & & \multicolumn{2}{c}{SlowFast} & \multicolumn{2}{c}{VideoST} & \multicolumn{2}{c}{SlowFast} & \multicolumn{2}{c}{VideoST} & \multicolumn{2}{c}{SlowFast} & \multicolumn{2}{c}{VideoST} \\
				\cmidrule(lr){5-6} \cmidrule(lr){7-8} \cmidrule(lr){9-10} \cmidrule(lr){11-12} \cmidrule(lr){13-14} \cmidrule(lr){15-16}
				
				& & & & Top1 & Top5 & Top1 & Top5 & Top1 & Top5 & Top1 & Top5 & Top1 & Top5 & Top1 & Top5 \\
				\midrule[0.5pt]
				
				\checkmark & \checkmark & \checkmark & \checkmark & \bf88.42 & \bf97.80 & \bf88.04 & \bf98.07 & \bf60.23 & \bf82.38 & \bf75.83 & \bf93.22 & \bf47.28 & \bf76.03 & \bf57.31& \bf86.87 \\
				& \checkmark & \checkmark &            & 86.93 & 96.38 & 86.27 & 96.63 & 58.31 & 81.84 & 74.76 & 90.70 & 47.04 & 75.67 & 56.75 & 83.78 \\ 
				\checkmark &            & \checkmark & \checkmark & \underline{87.39} & 97.18 & 87.23 & \underline{97.88} & \underline{58.91} & 82.15 & 75.34 & \underline{92.89} & \underline{47.11} & 75.87 & 57.05 & \underline{86.46} \\             
				\checkmark & \checkmark &            & \checkmark & 87.23 & \underline{97.37} & \underline{87.59} & 97.63 & 58.70 & \underline{82.22} & \underline{75.56} & 92.45 & 47.09 & \underline{75.92} & \underline{57.17} & 85.93 \\ 
				\checkmark & \checkmark & \checkmark &            & 86.95 & 96.58 & 86.53 & 97.59 & 58.34 & 81.92 & 74.92 & 92.38 & 47.04 & 75.72 & 56.83 & 85.84 \\
				&            &            &            & 84.82 & 95.73 & 85.80 & 96.48 & 55.23 & 80.27 & 72.31 & 90.83 & 45.31 & 73.82 & 54.27 & 83.82 \\
				\bottomrule[0.75pt]
			\end{tabular}
		}
	}
\end{table*}

\subsection{Discussion on class-imbalance issue} 
To examine the effectiveness of our method in the class-imbalance setting, we simulate the long-tail distribution characteristic. In particular, we randomly discard a portion of samples from the original balanced dataset, \eg, UCF101, such that the number of samples per category follows a power-law distribution $n_c = max(n_{\text{max}} \cdot c^{-\alpha},10), \quad c = 1, 2, \dots, C,$. Here, $n_{\text{max}}$ denotes the sample number of the most frequent class (\eg, $c=1$), and $\alpha > 0$ is the power-law exponent that controls the decay rate of sample sizes across categories. All categories are sorted in ascending order of their class index $c$, such that a larger $c$ corresponds to a category with fewer samples. Under this setting, the total number of training samples is reduced from 9,537 to 4,807, less than 50\% of the original size, while the sample distribution strictly adheres to the power law. 

The results are shown in Table~\ref{tbl:ucf101-imbalance}. From the table, both traditional KD methos and the state-of-the-art SAKD are constrained by the fixed sample distribution, resulting in limited performance gains. In contrast, our method achieves significant improvements across both benchmark architectures including SlowFast (CNN) and VideoST (Transformer). For example, our Top-1 accuracy reaches 76.07\% using SlowFast, even surpassing the teacher who gets 75.75\%. When using VideoST architecture, our method narrows the gap between student and teacher from the original 3.48\% to a mere 0.05\%. This provides the evidence that refining original samples in the frequency domain via adaptive sample generation compensates for the lack of features in tail categories, while channel-wise dynamic distillation helps to capture those important motion knowledge.

\begin{table}[!t]
	\centering
	\caption{Performance of the class-imbalance setting on UCF101 \cite{soomro-arxiv2012-ucf101}.}
	\label{tbl:ucf101-imbalance}
	\scalebox{0.9}{
		\setlength{\tabcolsep}{3mm}{ 
			\begin{tabular}{l l cc cc}
				\toprule[0.75pt]
				\multirow{2}{*}{Method} & 
				\multirow{2}{*}{Venue} & 
				\multicolumn{2}{c}{SlowFast\cite{feichtenhofer-iccv2019-slowfast}} & 
				\multicolumn{2}{c}{VideoST\cite{liu-cvpr2022-video}} \\
				\cmidrule(lr){3-4} \cmidrule(lr){5-6}
				& & Top1$\uparrow$  & Top5$\uparrow$  & Top1$\uparrow$  & Top5$\uparrow$ \\
				\midrule[0.5pt]
				Teac. & - & 75.75 & 91.15 & 74.82 & 90.28 \\
				Stud. & - & 72.56 & 89.58 & 71.34 & 88.27 \\
				\midrule[0.5pt]
				CKD \cite{shu-iccv2021-channelkd} & ICCV'21 & 73.54 & 90.41 & 72.95 & 89.52 \\
				DKD \cite{zhao-cvpr2022-dkd} & CVPR'22 & 73.81 & 90.55 & 73.12 & 89.70 \\
				CTKD \cite{li-aaai2023-ctkd} & AAAI'23 & 74.12 & 90.68 & 73.28 & 89.85 \\
				GKD \cite{wang-aaai2024-gkd} & AAAI'24 & 74.43 & 90.82 & 73.35 & 90.04 \\
				CrossKD \cite{wang-cvpr2024-crosskd} & CVPR'24 & 74.58 & 90.95 & 73.42 & 90.21 \\
				DualKD \cite{wang-tip2024-dkd} & TIP'24 & 74.72 & 91.08 & 73.49 & 90.58 \\
				DCSF \cite{dai-aaai2025-channelkd} & AAAI'25 & 74.85 & 91.19 & 73.52 & 90.75 \\
				SAKD\cite{li-mm2025-sakd} & MM'25 & \underline{75.04} & \underline{91.27} & 73.58 & \underline{90.92} \\
				DIST+ \cite{huang-tpami2025-dist} & TPAMI'25 & 74.98 & 91.22 & \underline{74.15} & 90.86 \\
				\midrule[0.5pt]
				\bf Ours & - & \bf 76.07 & \bf 91.48 & \bf 74.77 & \bf 91.03 \\
				\bottomrule[0.75pt]
			\end{tabular}
		}
	}
\end{table}

\subsection{Training Efficiency}
Since the inference time of student keeps the same, we examine the training time of different KD methods on both video and image datasets in Table~\ref{tbl:time}. From the table, our method achieves competitive training efficiency while maintaining strong performance, offering a favourable balance between computational cost and effectiveness. It is significantly faster than other recent methods such as GKD and DCSF, especially efficient on larger datasets, \eg, 3.12 hrs vs 3.74 hrs by GKD on Kinetics-400 using SlowFast model. Moreover, GKD and DCSF show the highest computational costs across most benchmarks; VideoST consistently takes longer to train than SlowFast across all methods; WRN requires more time than ResNet on CIFAR-100.

\begin{table*}[!ht]
	\centering
	\caption{Training efficiency comparison on five benchmarks.}
	\label{tbl:time}
	\scalebox{1.0}{
		\setlength{\tabcolsep}{1.2mm}{
			\begin{tabular}{l l cc cc cc cc cc}
				\toprule[0.75pt]
				\multirow{2}{*}{Method}& \multirow{2}{*}{Venue} & \multicolumn{2}{c}{UCF101 (min)\cite{soomro-arxiv2012-ucf101}} & \multicolumn{2}{c}{Kinetics-400 (hrs)\cite{kay-arXiv2017-kinetics}} 
				& \multicolumn{2}{c}{CIFAR-100 (sec)\cite{wah-2011-cifar}} & \multicolumn{2}{c}{ImageNet (hrs)\cite{krizhevsky-nips2012-imagenet}} 
				& \multicolumn{2}{c}{Sth-Sth-v2 (hrs)\cite{goyal-iccv2017-sthsth}} \\
				\cmidrule(lr){3-4} \cmidrule(lr){5-6} \cmidrule(lr){7-8} \cmidrule(lr){9-10} \cmidrule(lr){11-12}
				& & SlowFast & VideoST & SlowFast & VideoST 
				& ResNet & WRN & ResNet & MNv2
				& SlowFast & VideoST \\
				\midrule[0.5pt]
				FitNet\cite{romero-iclr2015-fitnets} &ICLR'15&9.12&25.02&2.58&2.74&37.23&42.28&1.84&2.32&2.34&3.24 \\
				GKD\cite{wang-aaai2024-gkd}  &AAAI'24&12.02&35.24&3.74&3.94&45.84&48.23&2.95&3.14&2.94&3.55 \\
				DualKD \cite{wang-tip2024-dkd}&TIP'24&10.92&29.24&3.56&3.67&41.73&46.42&2.34&2.93&2.85&3.35 \\
				DCSF	\cite{dai-aaai2025-channelkd} &AAAI'25&12.31&36.25&3.83&4.02&47.03&49.82&2.96&2.57&2.91&3.46 \\
				DIST+\cite{huang-tpami2025-dist}&TPAMI'25&11.53&35.29&3.45&3.56&44.92&47.28&2.86&3.12&2.67&3.52\\
				Ours && 11.42&30.11&3.12&3.16&39.41&44.17&2.05&2.48&2.46& 3.35\\
				\bottomrule[0.75pt]
			\end{tabular}
		}
	}
\end{table*}

\begin{figure}[!t]
	\centering
	\includegraphics[width=0.49\textwidth]{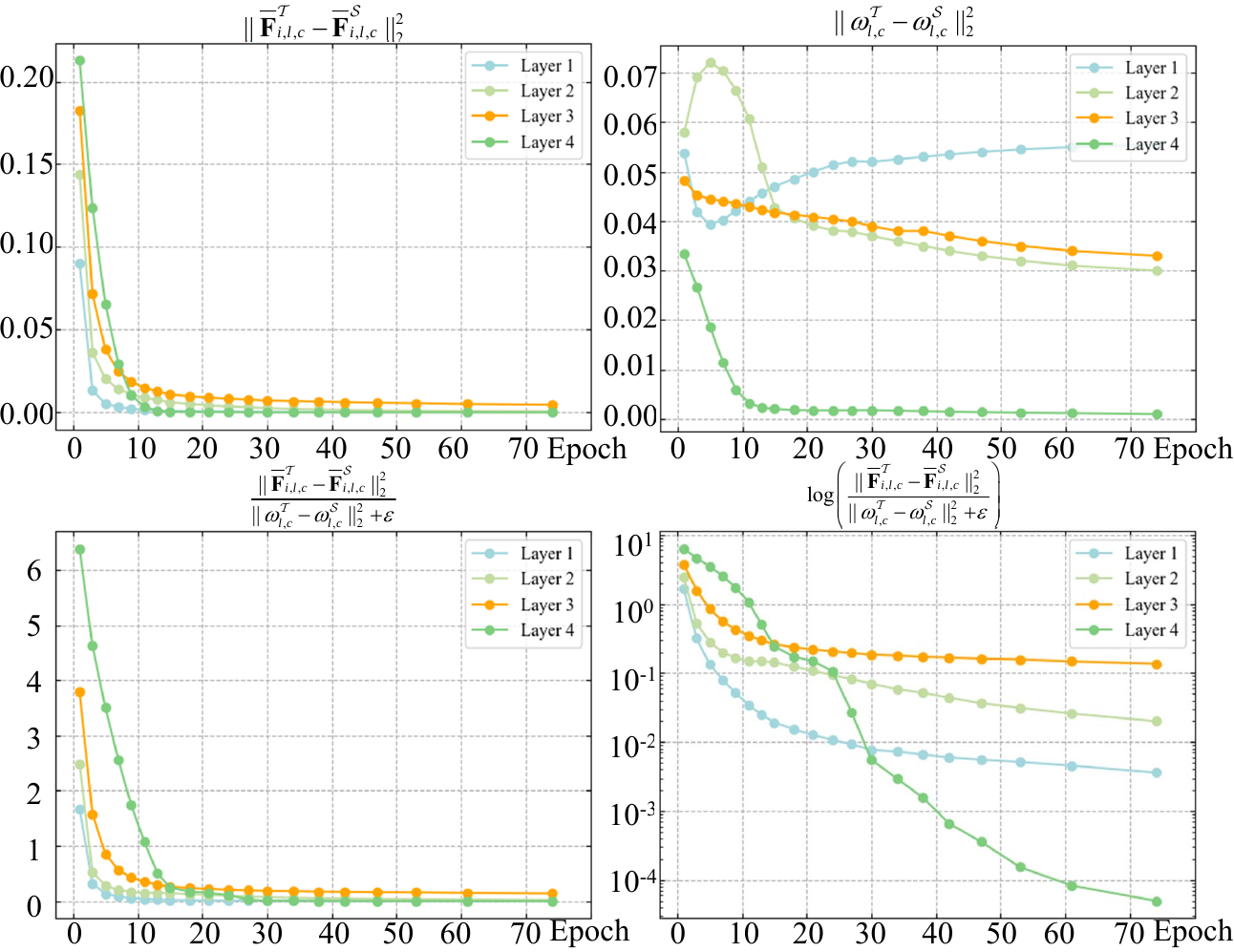}
	\caption{Convergence curves of different layers at stage one across epochs on UCF101 \cite{kay-arXiv2017-kinetics}.}
	\label{fig:curve}
\end{figure}

\subsection{Convergence Curve}
To display the trend of model convergence (weighted feature loss) at stage one during training, we draw the varying curves of different layers across epochs in Figure~\ref{fig:curve} for UCF101. As shown in the figures, we have the following observations: the feature difference drops rapid at early epochs and achieves stable after 20 epochs for all layers (\textit{top-left}); the centroid frequency difference has the similar trend but differ in layers (\textit{top-right}), but the former three layers get stable at very late epochs while the last layer only needs 15 epochs; when coupling the above two terms (\textit{bottom-left}), the weighted loss quickly converge after 15 epochs and low-level layers converges faster; the similar trend is also found when adopting the log scale for the weighted feature loss (\textit{bottom-right}).

\begin{figure}[!ht]
	\centering
	\begin{subfigure}{0.7\linewidth}
		\includegraphics[width=0.95\linewidth]{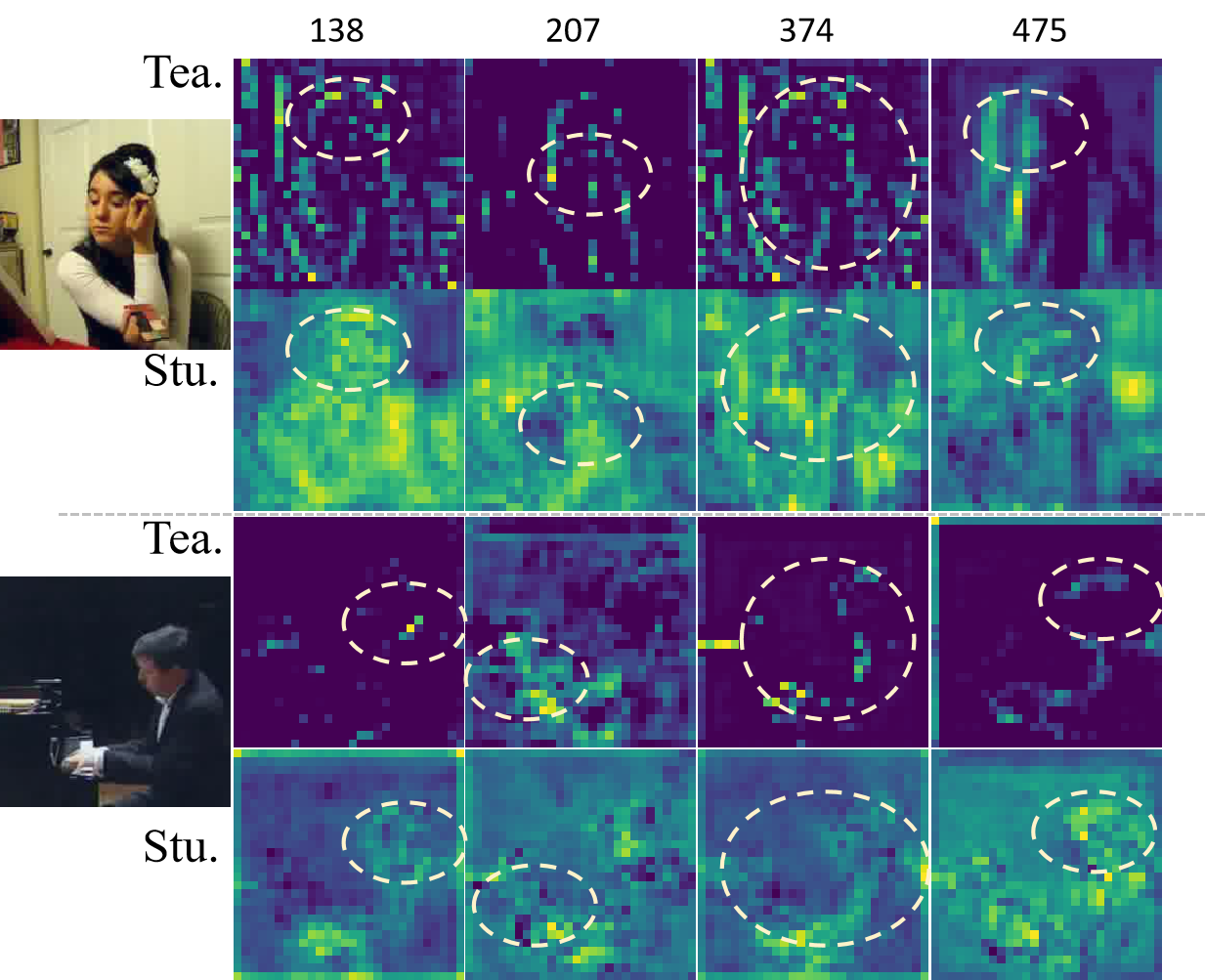}
		\caption{UCF101 \cite{soomro-arxiv2012-ucf101}.}
		\label{fig:channel_ucf}
	\end{subfigure}
	\begin{subfigure}{0.7\linewidth}
		\includegraphics[width=0.95\linewidth]{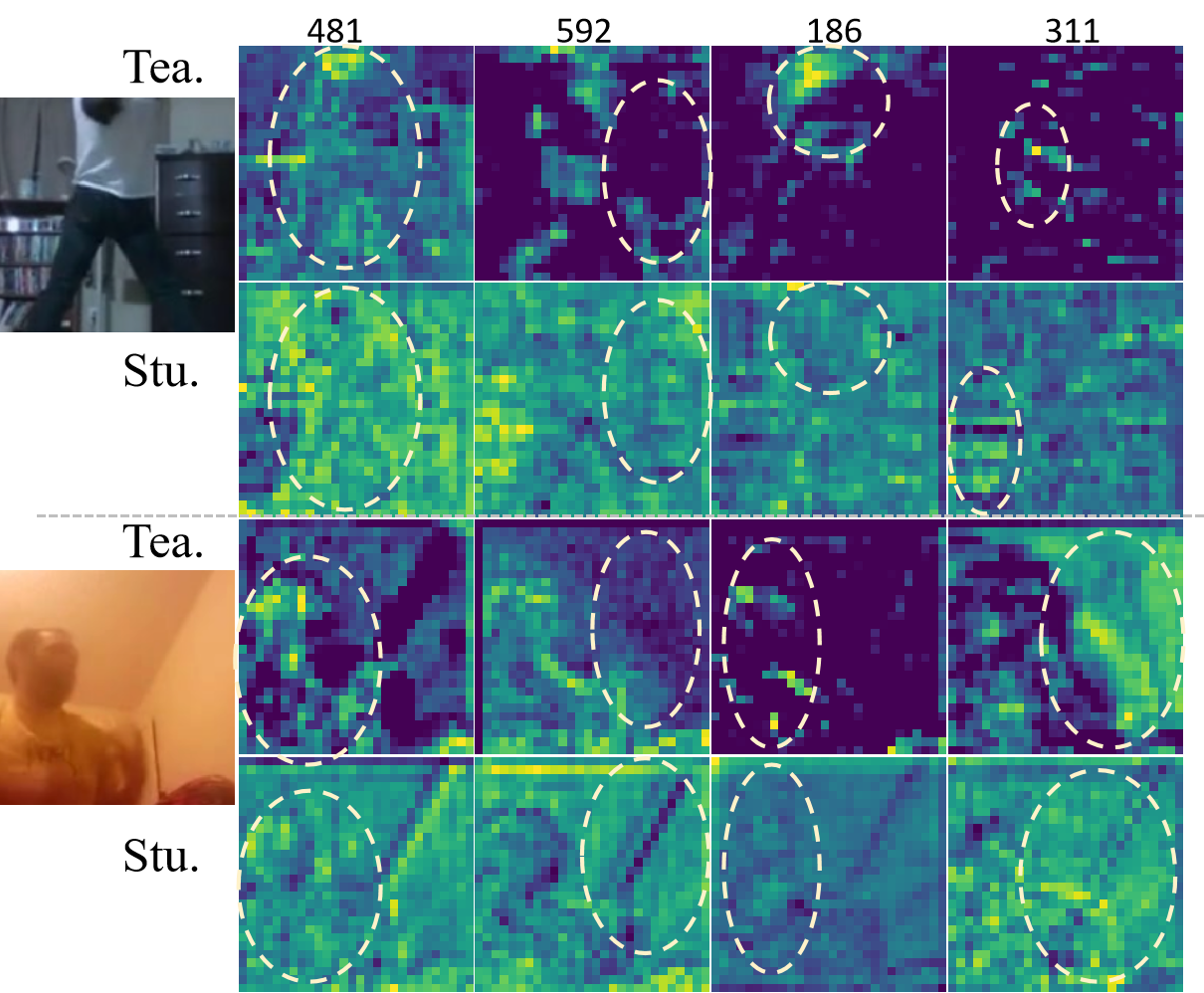}
		\caption{Kinetics-400 \cite{kay-arXiv2017-kinetics}.}
		\label{fig:channel_k400}
	\end{subfigure}
	\begin{subfigure}{0.7\linewidth}
		\includegraphics[width=0.95\linewidth]{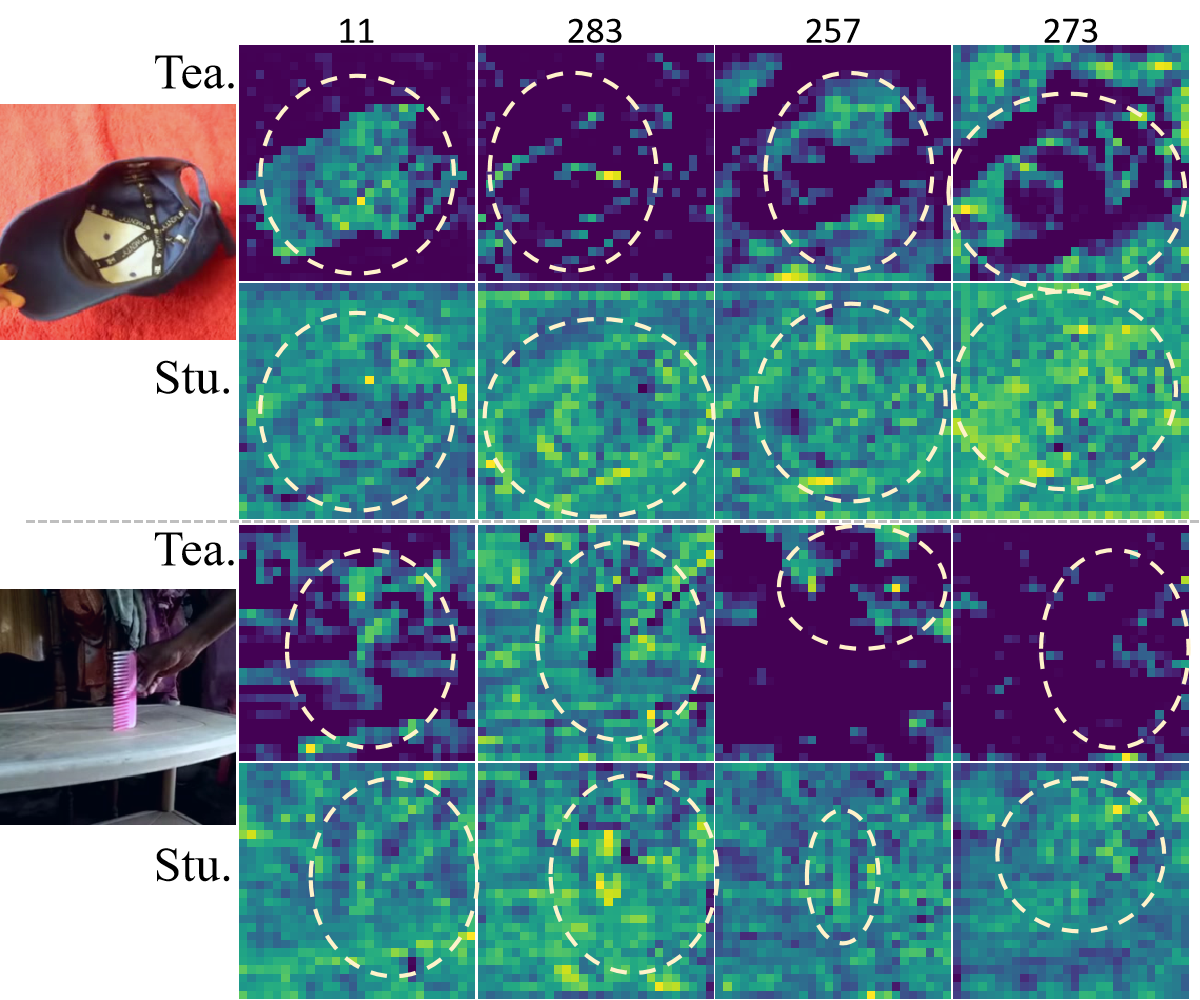}
		\caption{Sth-Sth-v2 \cite{goyal-iccv2017-sthsth}.}
		\label{fig:channel_sth}
	\end{subfigure}
	\vspace{2mm}
	\caption{Visualization of channel feature maps. (a) \textit{apply eye makeup} and \textit{playing piano}; (b) \textit{bending back} and \textit{contact juggling}; (c) \textit{turn something upside down} and \textit{fold something with hands}. }
	\label{fig:channel_video}
\end{figure}

\begin{figure}[!ht]
	\centering
	\begin{subfigure}{0.7\linewidth}
		\includegraphics[width=0.95\linewidth]{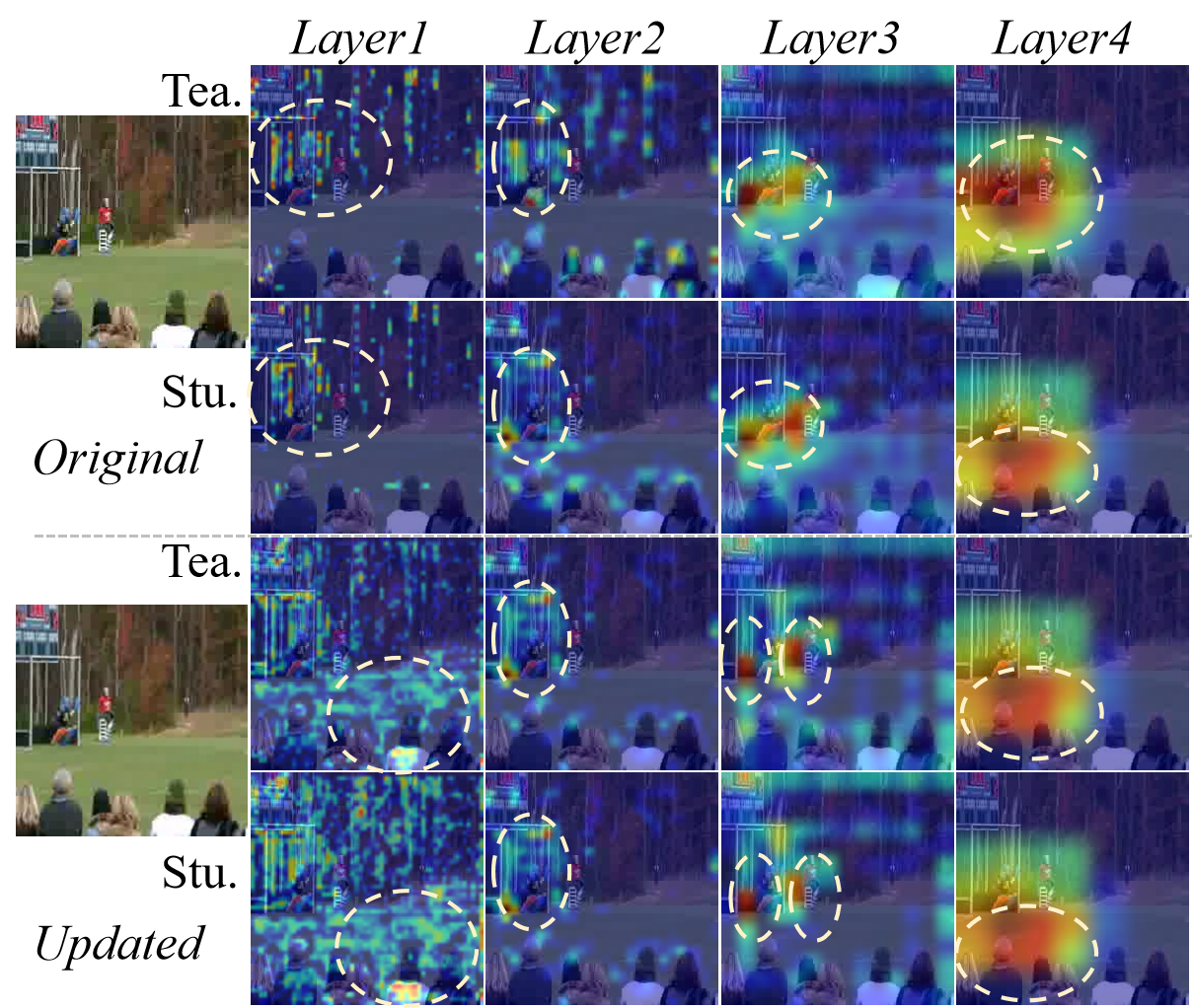}
		\caption{UCF101 \cite{soomro-arxiv2012-ucf101}.}
		\label{fig:fail_ucf}
	\end{subfigure}
	\begin{subfigure}{0.7\linewidth}
		\includegraphics[width=0.95\linewidth]{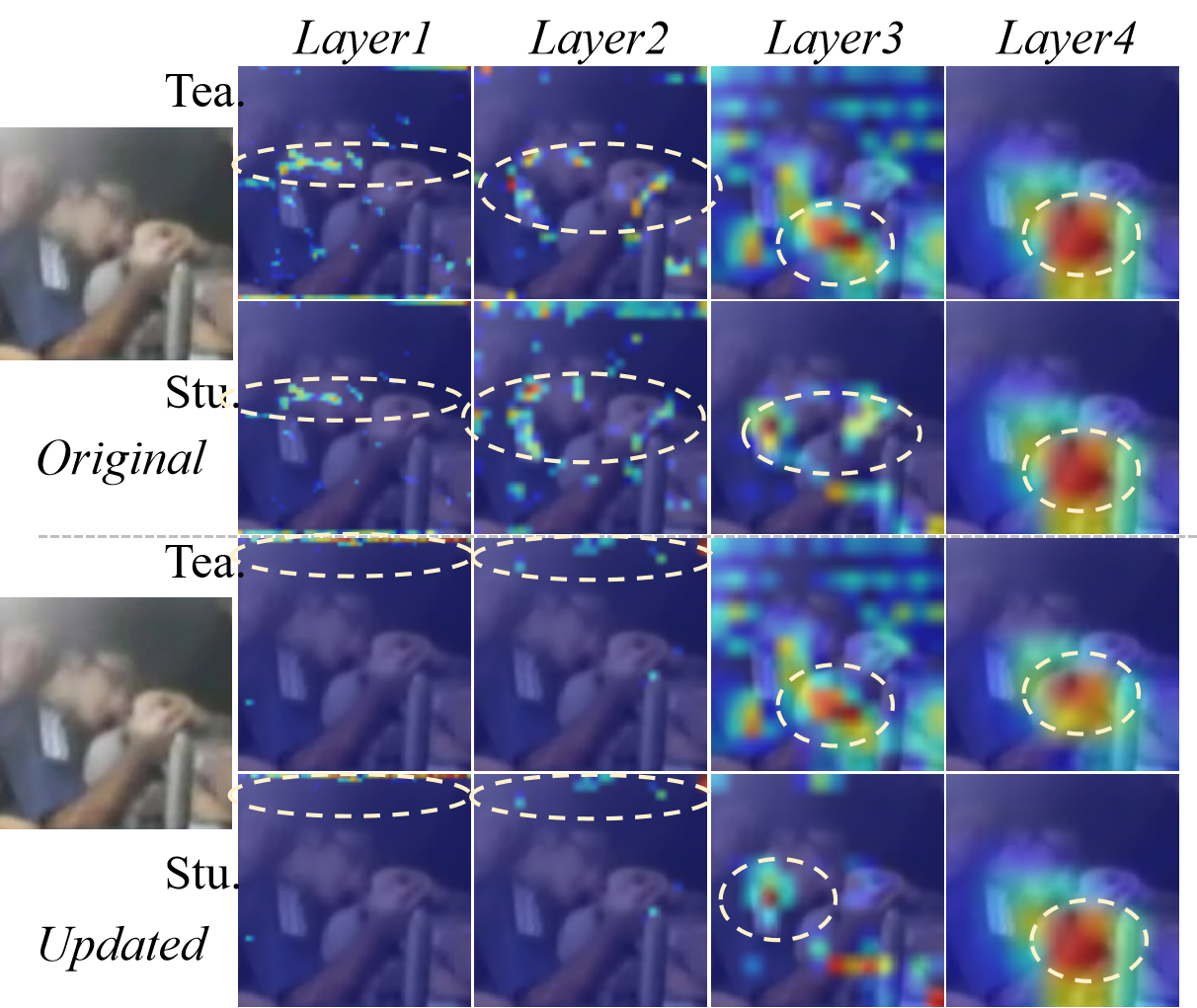}
		\caption{Kinetics-400 \cite{kay-arXiv2017-kinetics}.}
		\label{fig:fail_k400}
	\end{subfigure}
	\begin{subfigure}{0.7\linewidth}
		\includegraphics[width=0.95\linewidth]{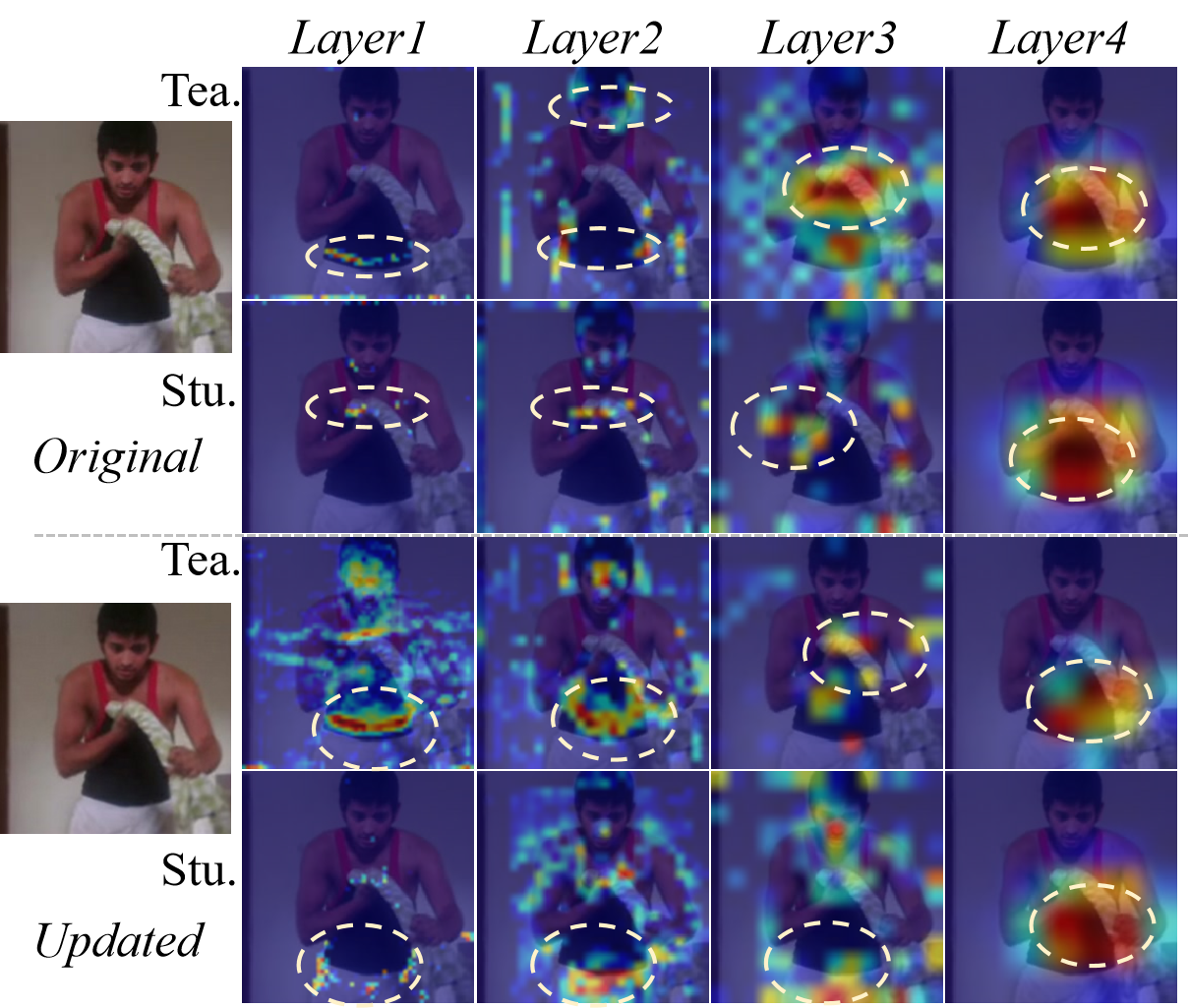}
		\caption{Sth-Sth-v2 \cite{goyal-iccv2017-sthsth}.}
		\label{fig:fail_sth}
	\end{subfigure} 	\vspace{2mm}
	\caption{Failure case. (a) \textit{field hockey penalty}; (b) \textit{arm wrestling}; (c) \textit{wring towel}. }
	\label{fig:fail_video}
	
\end{figure}

\subsection{Visualization of Channel Feature Maps}
Our method focuses on channel-wise knowledge distillation, and we visualize the channel feature maps to give an intuitive understanding in Figure~\ref{fig:channel_video}. We depict the feature maps of the top-4 channels with the largest centroid frequency difference at layer two for three video datasets, and show two examples each. 

Across all datasets, different channels exhibit significant semantic differences. In Figure~\ref{fig:channel_ucf}, channel 207 focuses on action-related features, such as hand movements in ``\textit{apply eye makeup}" (top) and ``\textit{playing piano}" (bottom), for both teacher and student; channel 374 shows a strong response to the person performing the action; channel 138 retains some background cues in addition to the action subject. In Figure~\ref{fig:channel_k400}, channel 481 primarily focuses on the moving person while maintaining some response to the background, such as wall edges (bottom); channel 311 pays more attention to background, \eg, the textural details of the bookshelf (top) and the geometric structure of the ceiling background (bottom).  In Figure~\ref{fig:channel_sth}, channel 11 captures the action-related objects such as the hat in the upper frame and the comb in the lower frame; channel 283 tends to extract contour and boundary of objects, showing sensitivity to structural features. These observations indicate that each channel may contribute to a diverse representation of the video content through distinct response patterns. This validates the rationale of designing the adaptive channel-wise KD framework.

\subsection{Failure Case}
To give an insight of the limitations of the proposed method, we show some failure cases of feature maps on three video datasets in Figure~\ref{fig:fail_video}.

From these figures, we have the following observations. In Figure~\ref{fig:fail_ucf}, it shows the hockey-throwing action where a large crowd in the background creates a significant distraction, student exhibits the attention drift in both shallow (Layer 1) and deep (Layer 4) features, tending to focus on the background crowd rather than the action. This might because the background regions contain rich mid- and high-frequency textures, which confuse with the edge cues of those persons in the background. In Figure~\ref{fig:fail_k400}, it displays the arm-wrestling action, where the two participants remain in a prolonged stalemate, resulting in minimal overall movement and weak temporal variation. This make our method struggle to extract discriminative frequency-varying signals along the temporal dimension, which weakens the intermediate-feature alignment between teacher and student. In Figure~\ref{fig:fail_sth}, it depicts the wring towel action, which shows subtle spatial variations along the temporal dimension. This suggests that we should consider the actions with weak temporal patterns by modeling local details. 

\end{document}